\pdfoutput=1 %
\documentclass{article} %
\usepackage[preprint]{icml2026}

\usepackage{hyperref}

\usepackage{float}
\PassOptionsToPackage{usenames,dvipsnames}{color}
\PassOptionsToPackage{usenames,dvipsnames}{xcolor}

\usepackage{booktabs}       %
\usepackage{nicefrac}       %
\usepackage{microtype}      %
\usepackage[flushleft]{threeparttable}

\usepackage{url}
\usepackage{xurl} %
\usepackage{xspace}
\usepackage{etoolbox}
\usepackage{fontawesome}
\usepackage{enumitem}
\usepackage{amsmath,amssymb}
\usepackage{array}
\usepackage{multirow}
\usepackage{wrapfig}
\usepackage{adjustbox,booktabs}
\usepackage{siunitx}
\usepackage{caption}
\usepackage[capitalize]{cleveref}
\usepackage{fontawesome}
\usepackage{stackengine}
\usepackage{svg}
\usepackage{mathtools}
\usepackage{xspace}
\usepackage{wrapfig}
\usepackage{makecell}
\usepackage{graphicx}
\usepackage{booktabs}
\usepackage{longtable}
\usepackage{pifont}
\usepackage{subcaption}
\usepackage[most]{tcolorbox}
\usepackage{changepage} %
\usepackage{colortbl}

\usepackage{tikz}
\usetikzlibrary{shapes.geometric}
\usetikzlibrary{arrows}
\usetikzlibrary{automata}
\usetikzlibrary{calc}
\usetikzlibrary{backgrounds}
\usetikzlibrary{decorations.markings}
\usetikzlibrary{decorations.pathmorphing}
\usetikzlibrary{decorations.pathreplacing}
\usetikzlibrary{fit}
\usetikzlibrary{patterns}
\usetikzlibrary{positioning}
\usetikzlibrary{shadows}
\usetikzlibrary{shapes}
\usetikzlibrary{arrows.meta}
\usetikzlibrary{shadows.blur}

\usepackage{amsthm}

\newcolumntype{x}[2]{S[table-format=#1.#2,table-auto-round]}

\definecolor{acceptblue}{HTML}{6494EA}
\hypersetup{
    colorlinks=true,
    linkcolor=acceptblue,
    citecolor=acceptblue,
    urlcolor=acceptblue,
}

\definecolor{lightred}{HTML}{ffcbc7}
\definecolor{gemini}{HTML}{4285F4}
\definecolor{claude}{HTML}{f3e9d7}
\definecolor{oai}{HTML}{10a37f}

\lstdefinestyle{mystyle}{
    breaklines=true,
    basicstyle=\scriptsize\ttfamily,
    numbers=none,
    language={},
    framextopmargin=0pt,
    framexbottommargin=0pt,
    breakindent=0pt,
    showspaces = false,
    keywordstyle=\bfseries,
    showstringspaces=false,
    columns=fullflexible,
    morekeywords={Answer}
    moredelim=[**][\bfseries]{!!}
}

\newtcblisting{prompt}[2][]{
    arc=3pt, outer arc=3pt,
    width=\linewidth,
    left=1mm,
    top=0mm,
    bottom=0mm,
    title=#2, 
    colback=gray!5!white,
    colframe=black!75!black,
    fonttitle=\bfseries,
    listing only, 
    listing options={style=mystyle,deletekeywords={}},
    breakable,
    #1
}

\newcommand{\gptfive}{\textsc{GPT-5}\xspace}
\newcommand{\gptfivefour}{\textsc{GPT-5.4}\xspace}

\newcommand{\gptfivetwo}{\textsc{GPT-5.2}\xspace}
\newcommand{\gptfivefive}{\textsc{GPT-5.5}\xspace}

\newcommand{\geminipro}{\textsc{Gemini-2.5-Pro}\xspace}
\newcommand{\claudeopus}{\textsc{Claude-Opus-4.1}\xspace}
\newcommand{\claudeopusfoursix}{\textsc{Claude-Opus-4.6}\xspace}

\newcommand{\grokfour}{\textsc{Grok-4}\xspace}

\newcommand{\geminithree}{\textsc{Gemini-3-Pro}\xspace}
\newcommand{\geminithreeone}{\textsc{Gemini-3.1-Pro}\xspace}

\newcommand{\ofour}{\textsc{o4-mini}\xspace}

\newcommand{\numberacceptedquestions}{77} %
\newcommand{\numberacceptedsubquestions}{129} %
\newcommand{\numberquestionsunderreview}{51} %
\newcommand{\numberquestionsdraft}{17} %
\newcommand{\numberirtquestions}{62} %
\newcommand{\numbergradedirtgptfivefour}{39} %
\newcommand{\numbergradedirtothers}{48} %
\newcommand{\numbersubquestionfigquestions}{40} %
\newcommand{\numberevaluatedmodels}{14} %
\newcommand{\numberparticipantswithacceptedquestion}{44} %
\newcommand{\numberopenproblems}{23} %
\newcommand{\correlationmainsub}{0.51} %

\newcommand{\agreementquestions}{18} %
\newcommand{\agreementcomparisons}{144} %
\newcommand{\progressexactagreement}{56.2} %
\newcommand{\progresswithinoneagreement}{89.6} %
\newcommand{\agreementhigh}{50.0} %
\newcommand{\agreementmoderate}{44.4} %
\newcommand{\agreementlow}{5.6} %
\newcommand{\agreementcorrectendresult}{93.7} %
\newcommand{\agreementhallucinatedfacts}{74.4} %
\newcommand{\agreementproblemunderstanding}{81.5} %
\newcommand{\agreementusefulprogress}{75.9} %
\newcommand{\agreementcalculationerror}{70.4} %
\newcommand{\agreementconceptualerror}{74.0} %
\newcommand{\agreementmathematicalinsight}{68.9} %
\newcommand{\agreementincorrectlogic}{66.7} %

\crefformat{section}{\S#2#1#3}
\crefrangeformat{section}{\S#3#1#4\crefrangeconjunction\S#5#2#6}
\crefmultiformat{section}{\S#2#1#3}{\crefpairconjunction\S#2#1#3}{\crefmiddleconjunction\S#2#1#3}{\creflastconjunction\S#2#1#3}
\newcommand{\crefrangeconjunction}{--}
\crefname{listing}{Lst.}{listings}
\crefname{line}{Lin.}{Lin.}
\crefname{appendix}{App.}{App.}
\Crefname{appendix}{App.}{App.}
\crefformat{appendix}{App.~#2#1#3}
\Crefformat{appendix}{App.~#2#1#3}

\icmltitlerunning{IMProofBench: Benchmarking AI on Research-Level Mathematical Proof Generation}

\begin{document}
\sisetup{
text-series-to-math = true,
propagate-math-font = true
}
\twocolumn[
    \icmltitle{IMProofBench: Benchmarking AI on Research-Level \\ Mathematical Proof Generation}

    \begin{icmlauthorlist}
        \icmlauthor{Johannes Schmitt}{yyy}
        \icmlauthor{Gergely Bérczi}{zzz}
        \icmlauthor{Jasper Dekoninck}{yyy}
        \icmlauthor{Jeremy Feusi}{yyy}
        \icmlauthor{Tim Gehrunger}{yyy}
        \par{\leftskip=0pt \rightskip=0pt \parfillskip=0pt plus 1fil
        \textbf{Benchmark Question Authors:} Raphael Appenzeller,
Pieter Belmans,
Alessio Bottini,
Jim Bryan,
João Camarneiro,
Ana Cannas da Silva,
Niklas Canova,
Ana-Maria Castravet,
Timo de Wolff,
Claudio Fontanari,
Filippo Gaia,
Baran Hashemi,
Daniel Holmes,
David Holmes,
Aitor Iribar Lopez,
Victor Jaeck,
Martina Jørgensen,
Steven Kelk,
Martijn Kool,
Stefan Kuhlmann,
Adam Kurpisz,
Johannes Lengler,
Chiara Meroni,
Ingmar Metzler,
Martin Möller,
Samuel Muñoz-Echániz,
David Muñoz-Lahoz,
Robert Nowak,
Georg Oberdieck,
Daniel Platt,
Dylan Possamaï,
Gabriel Ribeiro,
Aluna Rizzoli,
Daria Sakhanda,
Raúl Sánchez Galán,
Zheming Sun,
Diaaeldin Taha,
Josef Teichmann,
Richard P. Thomas,
Henk van der Pol,
Michel van Garrel,
Charles Vial\par
        \textbf{Benchmark Question Reviewers:}
Ignacio Barros,
Benjamin Doerr,
Peter Grünwald,
Henry Liu,
David Martins,
Aleksandar Mijatović,
Sergej Monavari,
Marc Roth,
Patrick Schnider,
Yannik Schuler,
Pim Spelier,
Yuuji Tanaka,
Ronald van Luijk\par}
    \vspace{2mm}
    \begin{center}
        \raisebox{-0.2\height}{\includegraphics[height=1em]{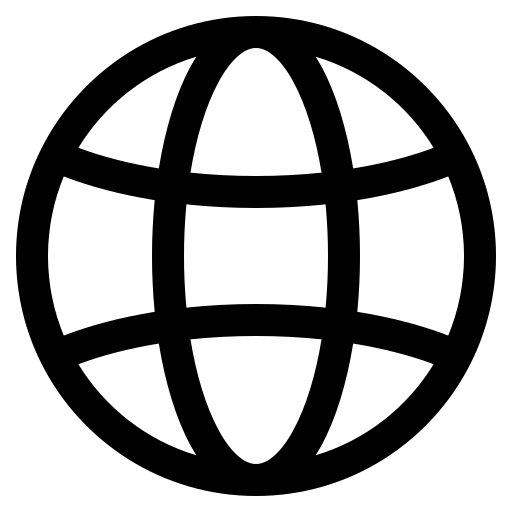}} \url{https://improofbench.math.ethz.ch}
        \end{center}
    \end{icmlauthorlist}

    \icmlaffiliation{yyy}{ETH Zurich}
    \icmlaffiliation{zzz}{Aarhus University}

    \icmlcorrespondingauthor{Johannes Schmitt}{johannes.schmitt@math.ethz.ch}

    \icmlkeywords{Machine Learning, ICML}

    \vskip 0.3in
]

\printAffiliationsAndNotice{}

\vspace{-5mm}

\begin{abstract}
As the mathematical capabilities of large language models (LLMs) improve, it becomes increasingly important to evaluate their performance on research-level tasks at the frontier of mathematical knowledge.
However, existing benchmarks are limited, as they focus solely on final-answer questions or high-school competition problems.
To address this gap, we introduce IMProofBench, a private benchmark consisting of $\numberacceptedquestions$ peer-reviewed problems developed by expert mathematicians.
Each problem requires a detailed proof and is paired with subproblems that have final answers, supporting both an evaluation by human experts and a large-scale quantitative analysis through automated grading. Furthermore, unlike prior benchmarks, the evaluation setup simulates a realistic research environment: models operate in an agentic framework with tools  like web search for literature review and mathematical software such as SageMath.
Our results show that current LLMs can already solve a significant percentage of research-level questions. 
IMProofBench will continue to evolve as a dynamic benchmark in collaboration with the mathematical community, ensuring its relevance for evaluating the next generation of LLMs.
\vspace{-3mm}

\end{abstract}

\section{Introduction}
Large language models (LLMs) are making rapid progress on mathematical tasks, achieving strong results on challenging benchmarks like AIME \citep{matharena} and FrontierMath \citep{frontiermath}. In fact, LLMs are now increasingly used as mathematical collaborators \citep{ghrist2024latticevaluedbottleneckduality,raz2025euleraiunifyingformulas} and have even solved several open problems in mathematics \citep{dobriban2025solving}. However, to accurately quantify how current systems can contribute in such settings, benchmarks are needed that test capabilities at the frontier of mathematical research.

\vspace{-1.33mm}
\paragraph{Limitations of existing benchmarks}  Existing benchmarks fall short of this objective: most focus on high-school or university-level mathematics \citep{matharena,friedereval}, due to the difficulty associated with designing rigorous, research-level problems. The few benchmarks that do target more advanced mathematics, like FrontierMath \citep{frontiermath} and HLE \citep{lastexam}, focus exclusively on final-answer problems. As a result, they overlook proof-writing capabilities and allow models to apply shortcuts to reach the correct final answer without fully solving the problem \citep{epochai_frontiermath_tier4_2025}.

\begin{figure*}[t]
    \centering
        \input{figures/overview}
    \caption{Example IMProofBench problem. Models are tested on research-level questions in an agentic framework with tool access. Grading of the main reasoning is done by a human expert, while follow-up subquestions are evaluated using an automated parser. Of the four models discussed in \cref{app:sampleProb}, only \gptfive and \grokfour made more than minor progress on this question. For full details, see \cref{app:sampleProb}.}
    \label{fig:overview}
    \vspace{-3mm}
\end{figure*}
\vspace{-1.33mm}
\paragraph{This work: IMProofBench}
To fill this gap, we introduce IMProofBench, a private benchmark developed in collaboration with the mathematical research community to evaluate LLMs on research-level proof writing. IMProofBench is supported by initiatives that actively involve professional mathematicians and includes tasks ranging from challenging oral exam questions in a graduate course to open research questions based on the contributors' own work. Unlike static benchmarks, IMProofBench is designed as a platform for continuous evaluation: problems are added on a rolling basis, ensuring its continued relevance for evaluating the next generation of LLMs. Currently, IMProofBench consists of $\numberacceptedquestions$ problems developed in collaboration with over $\numberparticipantswithacceptedquestion$ mathematicians, with $\numberquestionsunderreview$ more questions in the latest stages of the problem creation pipeline.

\vspace{-1.33mm}
\paragraph{Problem creation pipeline} Each problem in IMProofBench is authored by a research mathematician within their area of expertise. Submissions undergo a rigorous review process by a core team member and an additional mathematician with expertise in the relevant field. Reviewers provide feedback that allows authors to refine their problems before finalization. Alongside the main proof-writing tasks, authors are encouraged to add follow-up subquestions with final answers that can be automatically graded. These follow-ups enable a comparison between proof-writing and final-answer performance, while also supporting lower-cost evaluation across a broader range of models.

\vspace{-1mm}
\paragraph{Evaluation process}
Evaluation is conducted in an agentic framework designed to mirror a research environment. Models have access to computational tools such as Python, SageMath \citep{sagemath}, and web search. Each model is first tasked with solving the main problem, followed by the associated follow-up subquestions. The main solution is graded by the problem's author, who assigns a score from zero to three. Graders also annotate the types of errors, such as logical mistakes, and identify specific areas of partial progress, such as correct intermediate insights. Follow-up answers are automatically evaluated by comparing them with ground-truth solutions. We illustrate this process in \cref{fig:overview}, which shows a sample problem with two model solutions.

\vspace{-1mm}
\paragraph{Key results}
We evaluate 10 LLMs on IMProofBench, with many more models being evaluated solely on the final-answer-based subset. The results indicate that  LLMs can already solve a small but meaningful fraction of research-level problems. Among those evaluated on the proof-based portion, \gptfivefour performs best, with an estimated $49\%$ of tasks solved completely, closely followed by \geminithreeone at $46\%$. Other models perform substantially worse, with \gptfivetwo and \claudeopusfoursix providing complete solutions for only $32\%$ of tasks. On the final-answer data, \gptfivefour also scores best with a score of $75\%$.

\vspace{-1mm}
\paragraph{Qualitative analysis}
Beyond aggregate scores, our analysis reveals that many models are prone to reasoning errors, ranging from simple logical mistakes to deep misconceptions unlikely to be exhibited by any professional mathematician in the relevant area. Indeed, $10\%$ of model solutions contain arguments revealing fundamental misunderstandings of mathematical concepts, as judged by human graders. Moreover, models frequently hallucinate existing results to obtain a (flawed) answer.
Finally, models almost never abstain from providing a solution attempt, preferring to present convincing but incorrect proofs rather than admit they are stuck.
At the same time, they also show a wide-ranging familiarity with existing literature and can often provide insights that could meaningfully support mathematicians on a substantial number of problems.

\vspace{-1mm}
\paragraph{Key contributions} 
Our key contributions are:
\vspace{-3mm}
\begin{itemize}[left=0.1em,itemsep=0.1em]
\item IMProofBench, a private and evolving benchmark for research-level problems, developed in collaboration with the mathematical community.
\item A systematic analysis of proof generation capabilities across state-of-the-art LLMs, as judged by human experts.
\item A qualitative analysis discussing the difficulties and strengths of current state-of-the-art models and their potential application to research-level mathematics.
\end{itemize}

\section{Related Work}\label{sec:related-work}

\paragraph{High school and undergraduate benchmarks}
High school and undergraduate problems are the most common source of mathematical benchmarks due to their wide availability. Examples include GSM8k \citep{gsm8k} and MATH \citep{hendrycks2021math}, along with more recent efforts such as OmniMath \citep{omnimath}, UGMathBench \citep{xu2025ugmathbench}, and MathArena \citep{matharena}. However, these benchmarks fail to measure model performance on realistic, research-level tasks. Furthermore, even the most challenging competition problems are increasingly tractable for state-of-the-art LLMs \citep{matharena}, meaning that these benchmarks are saturated.

\paragraph{Research-level benchmarks}
To move beyond competition problems, several benchmarks aim to capture research-level mathematical reasoning, though each has notable limitations, and none provide systematic proof evaluation. FrontierMath \citep{frontiermath} offers extremely challenging private problems, though privileged access by OpenAI raises concerns about evaluation fairness \citep{epochopenai}. Humanity’s Last Exam \citep{lastexam} crowd-sources expert-level questions across domains, including mathematics, but suffers from contamination risks due to its open nature and reports of substantial noise in the benchmark \citep{hlenoise}. RealMath \citep{realmath} sources problems from arXiv papers, enabling dynamic evaluation of research-level problems, but it is currently not being maintained. Finally, the UQ-Dataset \citep{uq_dataset} collects unsolved StackExchange questions, many of them mathematical. While promising, it lacks a systematic human evaluation of proof validity.

\paragraph{Proof-based benchmarking efforts}
The importance of evaluating proof-generation capabilities has recently gained attention, leading to a range of benchmarking efforts. For example, \citet{mahdavi2025brains} showed that models trained with reinforcement-style methods such as GRPO \citep{grpo} perform poorly at proof writing. However, more recent evaluations on the USAMO and IMO 2025 demonstrated substantial progress in the ability of frontier models to construct rigorous mathematical arguments \citep{usamo,matharena}. At the same time, other studies highlighted a persistent gap between final-answer accuracy and genuine proof-writing ability, indicating that final-answer benchmarks are not sufficient to measure mathematical capabilities \citep{guo2025right,openproofcorpus}. Several benchmarks now also use LLM-judges as evaluators \citep{imobench,ineqmath}.
Despite these advances, benchmarks remain focused on high school and undergraduate mathematics.

\paragraph{Formal math benchmarks}
A complementary line of work evaluates LLMs on their ability to generate proofs in formal systems such as Lean \citep{lean4}. Success in this setting typically requires fine-tuning frontier models for this particular task \citep{Deepseekproverv2,goedelproverv2}, as off-the-shelf LLMs perform poorly. Formal proofs offer the advantage of automatic verification and scalable evaluation. Benchmarks in this space include PutnamBench \citep{putnambench} and MiniF2F \citep{minif2f}, which formalize problems from well-known mathematics competitions into Lean or Isabelle.
Work on a Lean-based benchmark with research-level problems is currently in progress with the \emph{ProofBench} initiative \citep{proofbench}.

\section{Benchmark Methodology}\label{sec:improofbench}
In this section, we present the creation and evaluation process of IMProofBench. We begin by outlining our community outreach efforts (\cref{subsec:community}), followed by a description of the problem creation pipeline (\cref{subsec:question_creation}) and the evaluation methodology (\cref{subsec:model_eval}). Finally, we discuss the current state of the benchmark and our plans to maintain and extend it as a platform for continuous evaluation (\cref{subsec:improofbench}).

\subsection{Community Outreach}\label{subsec:community}

Creating a novel and diverse collection of research-level problems is a challenging task, requiring professional mathematicians from a wide range of fields. To facilitate this, we undertook several initiatives to engage the community:
\vspace{-3mm}
\begin{itemize}[left=0.1em,itemsep=0.1em]
\item \textbf{Workshops}: We organized several problem-creation sessions as satellite events at mathematical conferences.
\item \textbf{Posters and flyers}: We distributed informational materials in math common rooms and conference venues to reach graduate students, postdocs, and faculty.
\item \textbf{Personal outreach}: Organizers and motivated contributors invited their academic networks to participate.
\end{itemize}
\vspace{-2mm}
These efforts are ongoing as we continue to expand the benchmark. Informal surveys of contributors indicate that key motivating factors to participate include convenient access to frontier models via the platform, curiosity about AI-generated responses to submitted questions, and the opportunity for co-authorship on resulting publications for contributors whose questions are accepted.

\subsection{Problem Creation Pipeline}\label{subsec:question_creation}

\begin{figure*}[!t]
    \centering
    \resizebox{0.9\linewidth}{!}{%
        \input{figures/creation_workflow}
    }
    \vspace{-1mm}
    \caption{Workflow for question creation with peer review. A problem is only accepted once the reviewers have no further comments.}
    
    \label{fig:creation}
    \vspace{-3mm}
\end{figure*}

\paragraph{Question creation}
As shown in \cref{fig:creation}, authors draft questions through a dedicated web interface and can immediately test them on an instance of \gptfivefive configured with high reasoning effort, built-in web search and code interpreter tools, and safeguards such as a cap of 20 evaluations per day to prevent abuse. This LLM interaction allows quick, optional feedback on both difficulty and potential ambiguities. Importantly, problem selection criteria are independent of the model's performance on the draft question.
Where possible, authors are asked to include follow-up subquestions with unique, automatically gradable answers, with the option to assign point weights for the solution of different subquestions to reflect their importance. This facilitates broader evaluation of more models by reducing reliance on human grading. To guide contributions, authors receive detailed instructions that include illustrative examples and emphasize that questions should require PhD-level insight, while avoiding standard textbook exercises or computational problems. A complete description of the author instructions is provided in \cref{app:question_creation}.

\paragraph{Question peer-review process}
Once a question is submitted, an administrator recruits a reviewer whose expertise aligns with the problem's subject area. Reviewers are invited via email, with invitations extended to both existing benchmark participants and external experts if necessary.
The review process follows an academic peer-review model, with the administrator and reviewer providing detailed feedback, asking for revisions where necessary. While the reviewer concentrates on verifying mathematical correctness and difficulty, the administrator ensures that the submission adheres to the guidelines. Authors are then invited to revise their problem and respond to comments with clarifications or adjustments. A problem is accepted only after both the administrator and reviewer have no remaining concerns. A full description of the reviewer instructions is given in \cref{app:review_process}.

\subsection{Model Evaluation}\label{subsec:model_eval}

\paragraph{Evaluation environment}
As shown in \cref{fig:evaluation}, models are evaluated within an agentic framework designed to approximate real research conditions. We use the Inspect framework~\citep{inspect_ai} with these tools:
\vspace{-2mm}
\begin{itemize}[left=0.1em,itemsep=0em]
    \item \textbf{Python}: a full scientific environment with NumPy, SciPy, SymPy, and related libraries.
    \item \textbf{Bash}: an Arch Linux console with persistent filesystem and computer algebra systems like GAP \citep{gap}, and Maxima \citep{maxima}.
    \item \textbf{SageMath}: open-source mathematical software with specialized packages and databases \citep{sagemath}.
    \item \textbf{Web search}: a tool for literature search.
\end{itemize}
\vspace{-2mm}
A full description of these tools is provided in \cref{app:tools}. To submit an answer, models must use a submit tool, which ensures a clear distinction between reasoning steps and the final output. The submitted answer is either presented to the human grader for main questions or compared with ground-truth answers for follow-up subquestions. Each model is allocated up to 300,000 tokens for main questions, with an additional 100,000 tokens available for each follow-up. Due to a bug in our token accounting, reasoning tokens of OpenAI (and partially Anthropic) models were counted twice. This ended a few OpenAI runs early, at about 150k--185k true tokens ($5/39$ graded \gptfivefour answers, $11/48$ for \gptfivetwo, $1/61$ for \gptfive). All models use this framework except \textsc{GPT-5.2 Pro}, which was queried through the API with only its built-in web search. %

\begin{figure*}[t]
    \centering
    
    \resizebox{\linewidth}{!}{%
        \input{figures/evaluation_workflow}
    }
    \vspace{-6mm}
    \caption{Evaluation workflow in a multi-turn environment with research tools. The main solution is graded by a human expert, while follow-up questions are automatically evaluated.}
    \vspace{-3mm}
    \label{fig:evaluation}
\end{figure*}

\paragraph{Grading process} 
Scoring of model answers takes place in two separate stages. First, follow-up subquestions are automatically graded by comparing the model's output with the ground-truth reference. Currently, this automated evaluation is also manually verified by an administrator, who can correct parsing errors and update the grading script if necessary. In the second stage, human grading is conducted through our dedicated web interface. The question's author serves as grader and provides three types of feedback:
\vspace{-3mm}
\begin{itemize}[left=0.1em,itemsep=0.1em]
\item \textbf{Error classification}: identifying reasoning mistakes caused by incorrect logic, hallucinations, calculation errors, or conceptual misunderstandings.
\item \textbf{Achievement indicators}: marking whether the model showed understanding, reached correct conclusions, identified key insights, or produced useful reasoning.
\item \textbf{Overall progress}: assigning a score of no ($0/3$), minor ($1/3$), major ($2/3$), or full ($3/3$) progress.
\end{itemize}
\vspace{-2mm}
Error classification and achievement indicators are recorded as eight binary marks and enable a more fine-grained analysis of model performance. This structure allows us to identify both the areas where models can already assist research mathematicians and the areas where they remain most prone to errors. To avoid bias, model identities remain hidden until grading is completed.

In \cref{app:grading_reliability}, we report results from a small-scale analysis of grader reliability, in which a subset of model answers was graded multiple times by different judges. Graders assign exactly the same overall progress score in $\progressexactagreement\%$ of cases and agree within one point in $\progresswithinoneagreement\%$ of cases. Agreement across the binary categories ranges from $\agreementincorrectlogic\%$ to $\agreementcorrectendresult\%$ when ``Not Sure'' and ``N/A'' answers are excluded, reflecting the inherent subjectivity of these dimensions: what one mathematician considers useful or insightful may not be viewed the same way by another. Despite this variability, we believe these categories still provide valuable insight into how mathematicians view model outputs.

\subsection{Benchmark Statistics and Future Development} \label{subsec:improofbench}
\vspace{-1mm}
\paragraph{State of the benchmark}
IMProofBench is under active development, with this version consisting of $\numberacceptedquestions$ questions and $\numberacceptedsubquestions$ follow-up subquestions. Topics range from areas of pure mathematics, such as algebraic geometry, combinatorics, and graph theory, to applied subjects such as stochastic analysis and bioinformatics. In \cref{fig:wordcloud}  of \cref{app:details}, we include a word cloud of question tags, weighted by frequency.
Of the $\numberacceptedquestions$ benchmark problems, authors characterize $\numberopenproblems$ as open research questions.
A total of $\numberparticipantswithacceptedquestion$ mathematical researchers have contributed at least one question. %

\vspace{-1mm}
\paragraph{Continuous development} 
With models showing rapid progress in mathematics, benchmarks are being saturated at an accelerating pace. To ensure that IMProofBench remains both unsaturated and challenging, we are committed to its continuous development along several dimensions. First, we will maintain our problem creation pipeline and accept problems on a rolling basis, while forming new strategic partnerships with leading mathematical institutions to keep problem difficulty aligned with the capabilities of future models. Second, to prevent contamination, we will employ a dynamic problem management system in which authors are encouraged to revisit and possibly retire their problems once new publications or techniques make them easier. Third, we plan to create a transparent interface that allows major companies or research labs to use their own agents for solving problems in IMProofBench. This creates an opportunity to provide objective evaluations of the latest state-of-the-art models and agents in realistic settings. Other ideas for future work are given in \cref{app:development}.

\paragraph{Handling incomplete scores} As the benchmark grows and new models are evaluated, gaps will start appearing in the model scores on individual questions. For older questions, human annotators might not have the time anymore to grade the models, while older models will not be re-evaluated on new questions. Yet, we want to provide an accurate comparison between all models at any given time. For this purpose, we use an item response theory (IRT) model to fill in the gaps with the expected performance of a model on a given question, and compare models based on their expected performance. In particular, we fit a four-level Partial Credit Model (PCM) to the observed scores. The model assigns each model $m$ a latent ability $\theta_m$, each question $q$ a difficulty $b_q$, and each model a set of score-threshold parameters $\tau_m$ governing the transition between the score levels $0,1,2,3$. Concretely, the probability that model $m$ receives score $k$ on question $q$ is proportional to
\[
\Pr(X_{mq}=k) \propto \exp\!\left(\sum_{s=1}^{k} (\theta_m - b_q - \tau_{m,s})\right).
\]
After maximum likelihood estimation on the observed entries, each missing score is replaced by its model-implied expected value
\[
\mathbb{E}[X_{mq}] = \sum_{k=0}^{3} k \, \Pr(X_{mq}=k),
\]
and models are compared using their average expected score over all questions, and the expected value of completely correct questions. We fit the model on the $\numberirtquestions$ questions graded for at least four models. Confidence intervals are estimated using bootstrapping over the imputed scores. Resampling questions instead gives standard errors of about $\pm 7$ points for the five leading models.

\begin{figure}[t]
    \vspace{-3mm}
    \begin{center}
        \includegraphics[width=0.45\textwidth]{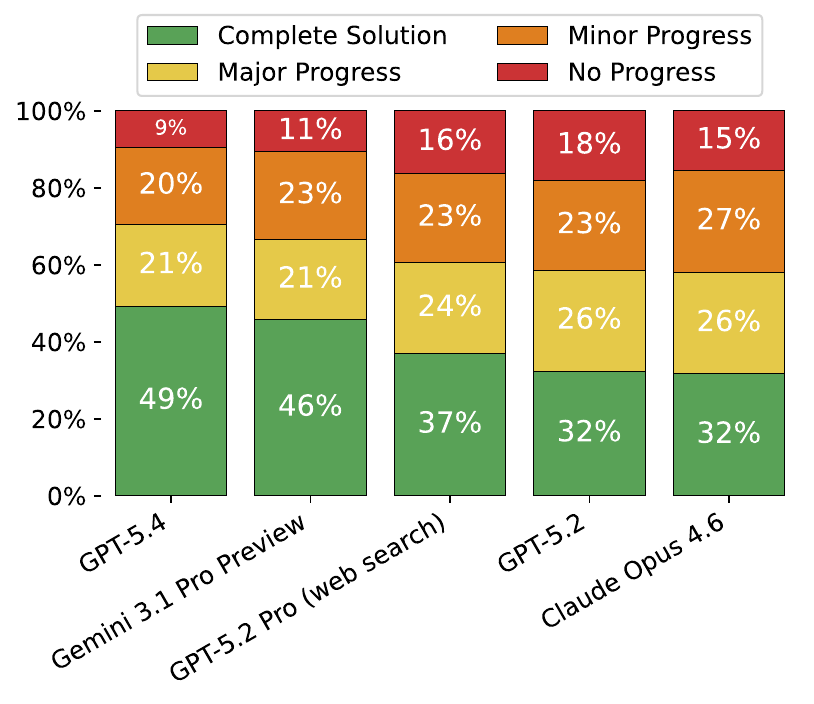}
    \end{center}
    \vspace{-6mm}
    \caption{Results on IMProofBench ($\numberirtquestions$ questions, partly imputed, see \cref{subsec:improofbench}). \textsc{GPT-5.2 Pro} was evaluated non-agentically.}
    \label{fig:results:proof}
    \vspace{-4mm}
\end{figure}
\begin{figure*}[t]
\centering
\includegraphics[width=\linewidth]{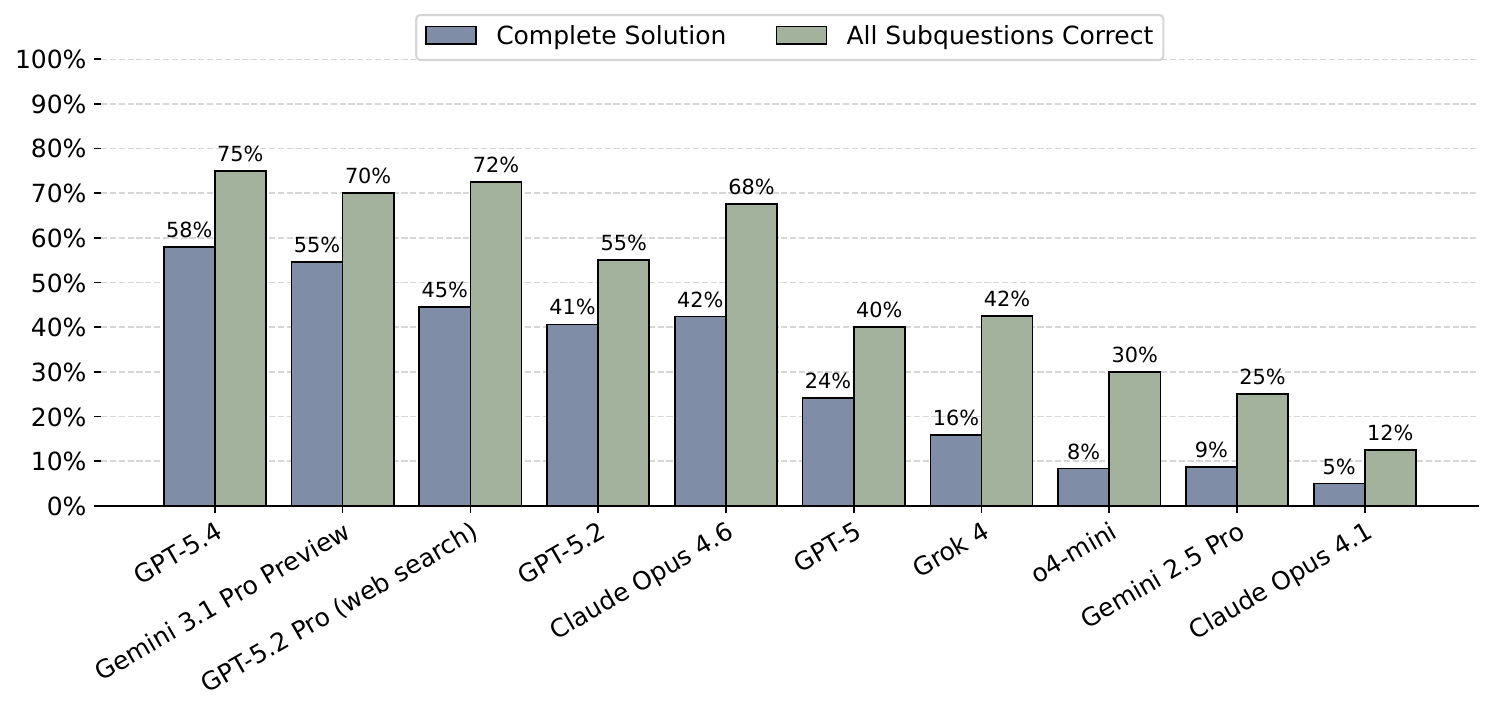}
\vspace{-8mm}
\caption{Results on the $\numbersubquestionfigquestions$ questions with follow-up subquestions. Complete-solution rates use IRT estimates for ungraded answers.} 
\vspace{-3mm}
\label{fig:main_results} 
\end{figure*}
\vspace{-2mm}
\section{Experimental Results}\label{sec:results}

In this section, we give quantitative and qualitative summaries of model performance on IMProofBench. In \cref{subsec:quantitative}, we compare final-answer correctness and proof-generation capabilities of several frontier LLMs. Then, in \cref{subsec:error_analysis}, we present a detailed analysis of errors and achievements made by these models. In \cref{sec:tools}, we analyze token and tool usage. We then perform an ablation on the agentic setup (\cref{sec:ablation}) and conclude in \cref{sec:behavior} with a qualitative discussion of several notable examples and overall results.

\subsection{Main Results}\label{subsec:quantitative}

\paragraph{Proof-based evaluation} As shown in \cref{fig:results:proof}, \gptfivefour achieves the strongest performance, producing a complete solution in an estimated $49\%$ of cases, although its lead over \geminithreeone is within the statistical uncertainty. It fails to make any progress on only $9\%$ of the questions, showing that the model can engage meaningfully with most problems in the benchmark. These results highlight the impressive capabilities of current systems. Among the $\numberopenproblems$ open problems, GPT-5 solved a first one in December 2025.

\begin{figure*}
  \centering
  \begin{minipage}[t]{0.48\textwidth}
    \includegraphics[width=\linewidth]{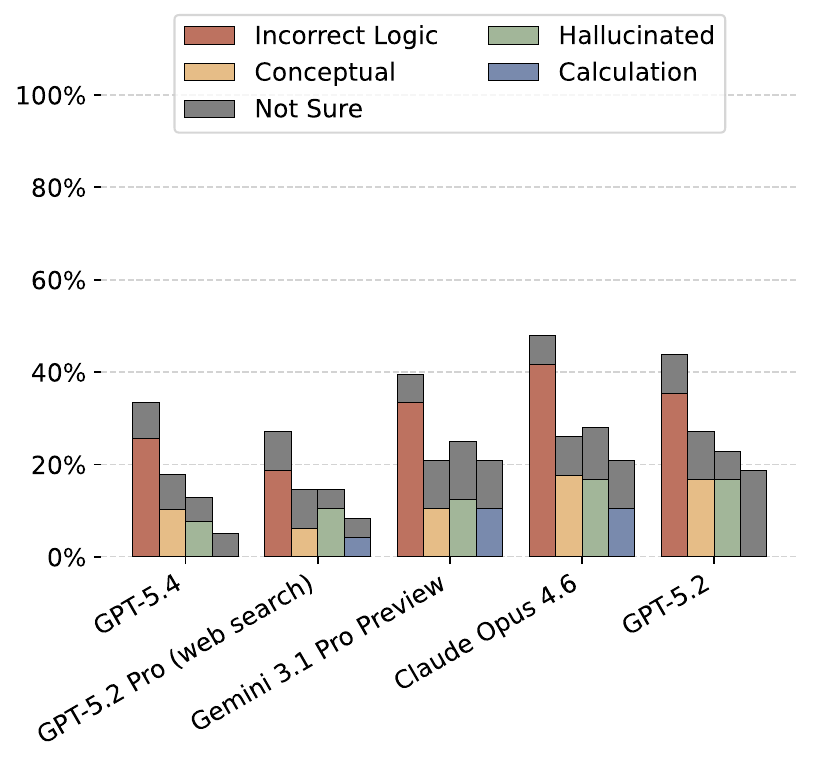}
    \vspace{-8mm}
    \caption{Error indicators (\% of human-graded answers).}
    \label{fig:errors:main}
  \end{minipage}\hfill
  \begin{minipage}[t]{0.48\textwidth}
    \includegraphics[width=\linewidth]{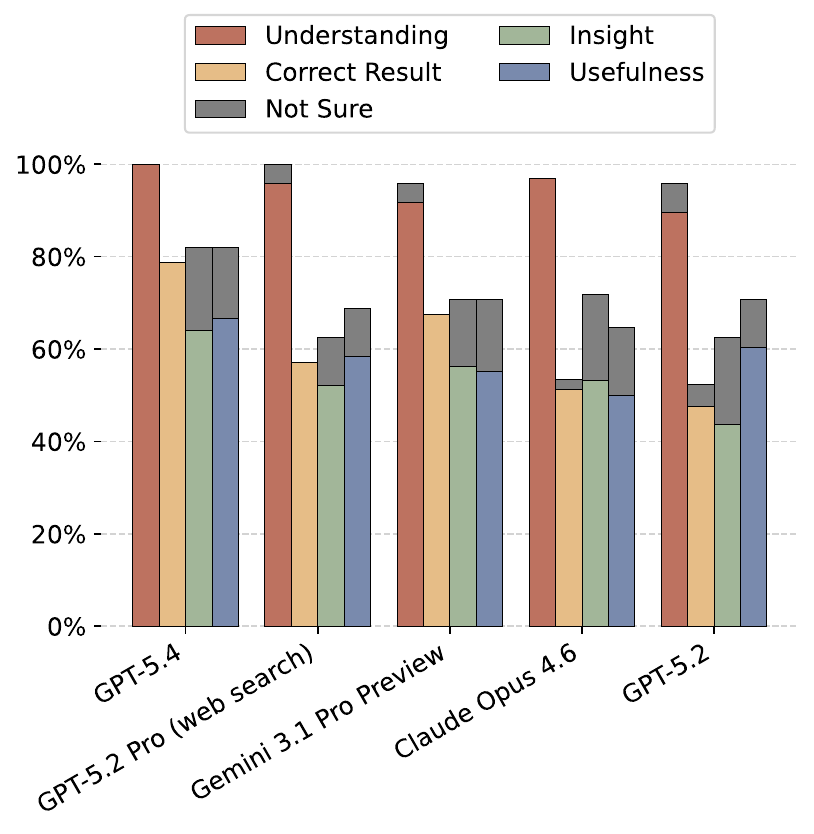}
    \vspace{-8mm}
    \caption{Achievement indicators (\% of human-graded answers).}
    \label{fig:achievements:main}
  \end{minipage}
  \vspace{-3mm}
\end{figure*}

\begin{figure*}[t]
  \centering
  \begin{minipage}[t]{0.47\textwidth}
    \includegraphics[width=\linewidth]{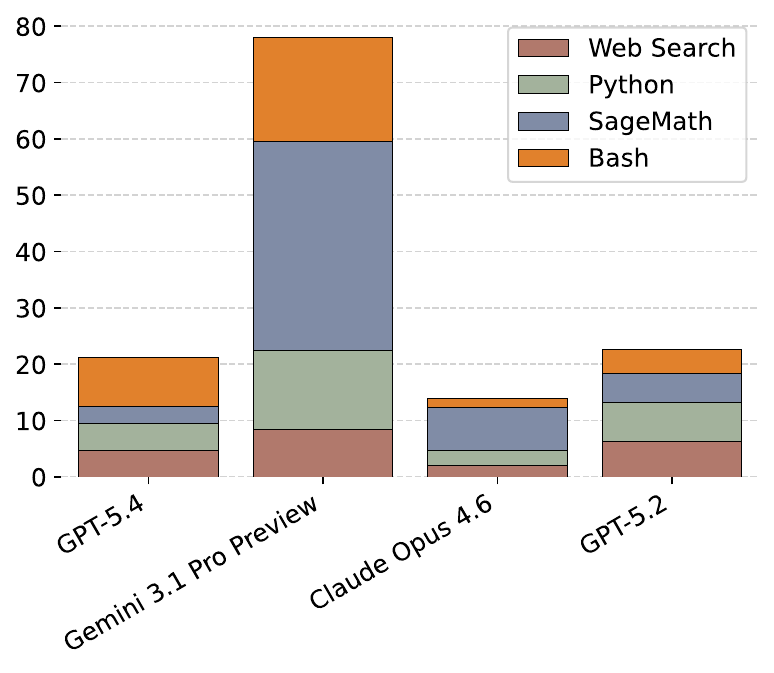}
    \vspace{-8mm}
    \captionof{figure}{Average tool usage per question.}
    \label{fig:tool:main}
  \end{minipage}\hfill
  \begin{minipage}[t]{0.47\textwidth}
    \includegraphics[width=\linewidth]{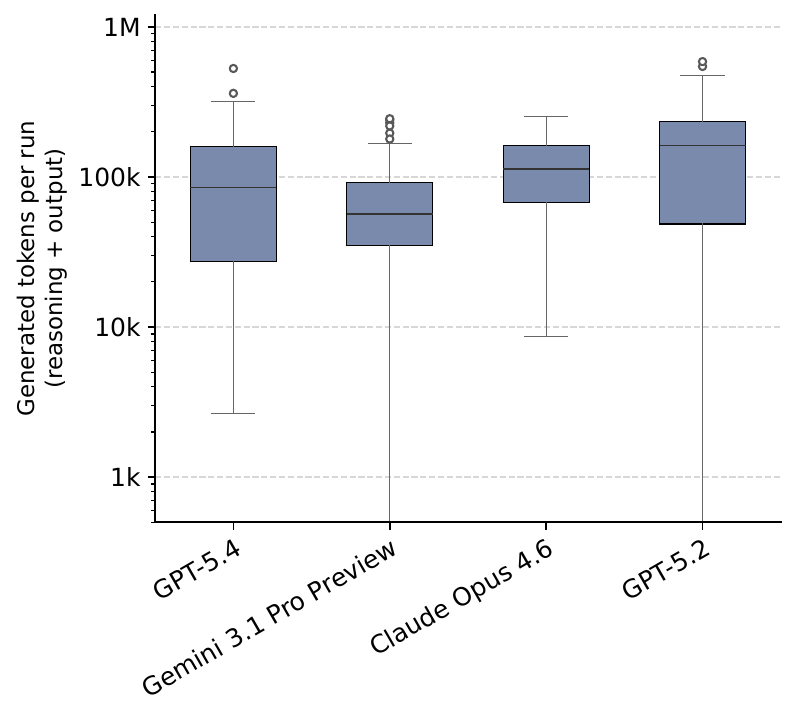}
    \vspace{-8mm}
    \caption{Token usage per question (reasoning + output).}
    \label{fig:tokens:main}
  \end{minipage}
  \vspace{-3mm}
\end{figure*}

\paragraph{Final-answer evaluation} In \cref{fig:main_results}, we compare performance between final-answer accuracy and full proof-based evaluation on the $\numbersubquestionfigquestions$ questions that include both. \gptfivefour also performs best on this final-answer portion, scoring $75\%$. The correlation coefficient between author-weighted subquestion scores and the $0$–$3$ progress scores assigned by human graders is $\correlationmainsub$. This suggests that final-answer evaluation is a useful proxy for model ability, but human grading provides essential nuance and a more refined view of performance. In \cref{app:details}, we analyze final-answer accuracy over the full set of questions, including partial progress.

\subsection{Error and Progress Analysis}\label{subsec:error_analysis}

We analyze the error and achievement indicators classified by the question authors ($\numbergradedirtgptfivefour$ graded answers for \gptfivefour, $\numbergradedirtothers$ for the others). This provides a clearer picture of where models fail and where they provide meaningful help.

\paragraph{Error indicators} As shown in \cref{fig:errors:main}, models make a variety of errors. Logical errors are the most common, with models frequently introducing unfounded assumptions or claiming incorrect implications. \claudeopusfoursix is particularly weak in this respect, having logical errors in $42\%$ of its responses. All other errors are much less common and remain below $20\%$. However, hallucinations still occur, which is a major issue for any scientific application.

\paragraph{Achievement indicators} As shown in \cref{fig:achievements:main}, most models demonstrate general familiarity with the background knowledge needed to understand the problems, which is an impressive achievement given that many of these questions reference highly specialized mathematical concepts. Creative ideas are rarer, but \gptfivefour still displays non-trivial creativity in over half of its solutions. This indicates that the model can already make remarkable progress on difficult problems. Finally, in some cases, models provide insights that could be helpful to expert mathematicians, with \gptfivefour offering meaningful contributions in about two thirds of its attempts. This is a significant achievement.

\subsection{Tool and Token Usage} \label{sec:tools}

As illustrated in \cref{fig:tool:main} and \cref{fig:tokens:main}, models vary widely in their resource usage, both in tool selection and token consumption. \geminithreeone uses about $3.5$ times as many tool calls as any other model, but at the same time also uses the fewest tokens. All models make use of all four tool types in varying proportions, which gives a strong indication that all tools are genuinely useful. Usage plots for the other models are shown in \cref{app:details}.

\subsection{Non-Agentic Ablation} \label{sec:ablation}
To evaluate the effectiveness of our agentic setup, we conduct an ablation study comparing model performance with and without the agentic setup on the final-answer portion of IMProofBench. For this ablation, the non-agentic evaluation mirrors the structure of the agentic setup, but without tool access and without a separate submit tool.

\textbf{Results} \cref{tab:ablation:main} reports the final-answer accuracy for both agentic and non-agentic settings across several models. The impact of the agentic setup varies substantially by model. While \grokfour, \geminithree, and \gptfivetwo benefit significantly from tool access, \gptfive only improves moderately, and \geminipro performs slightly worse.

To better understand this behavior, we perform a more fine-grained ablation on \gptfive by selectively enabling different tools, with results shown in \cref{tab:ablation:gptfive}. We observe that access to the web-search tool can yield gains up to $8.4\%$. In contrast, enabling the code interpreter can lead to a performance drop of $10\%$. We hypothesize that \gptfive tends to overuse the code interpreter, often treating mathematically-oriented problems as programming tasks when given access to this tool. This results in unnecessary use of the code tool that distracts from effective reasoning, degrading performance.

Since most models show strong improvements when equipped with these tools, these results suggest that the limitations do not come from our agentic framework, but rather from how individual models choose to leverage available tools. We therefore retain the full agentic configuration as our primary evaluation. This setting more closely reflects realistic mathematical problem-solving workflows, where external resources and computational tools are available.

\subsection{Qualitative Analysis} \label{sec:behavior}
We now describe qualitative observations drawn from manual inspection of logs and grader comments.

\begin{table}[t]
\centering
\begin{tabular}{lccc}
\toprule
Model & Agentic & Non-agentic & Diff \\
\midrule
\gptfive & $54.5\%$ & $51.9\%$ & $+2.6\%$  \\
\geminipro & $39.5\%$  & $40.4\%$& $-0.9\%$ \\
\grokfour & $57.5\%$  & $43.1\%$ &$+14.4\%$ \\
\gptfivetwo & $73.5\%$ & $63.1\%$ &$+10.4\%$ \\
\geminithree & $72.6\%$ & $63.1\%$ &$+9.5\%$ \\
\bottomrule
\end{tabular}
\vspace{1mm}
\caption{Agentic and non-agentic evaluations of several models on the final-answer portion of IMProofBench}
\vspace{-2mm}
\label{tab:ablation:main}
\end{table}

\begin{table}[t]
\centering
\begin{tabular}{c c cc}
\toprule
 & & \multicolumn{2}{c}{Code Interpreter} \\
\cmidrule(lr){3-4}
Full Agent & Web Search & No & Yes \\
\midrule
\multirow{2}{*}{$54.5\%$} & No & $51.9\%$ & $50.3\%$ \\
 & Yes & $60.3\%$ & $50.4\%$ \\
\bottomrule
\end{tabular}
\vspace{1mm}
\caption{Evaluation of \gptfive under several settings. The full agent differs from the other settings in that it also allows access to the bash tool, SageMath, and makes use of the submit tool.}
\vspace{-5mm}
\label{tab:ablation:gptfive}
\end{table}
\paragraph{Broad and deep literature knowledge} Leading models such as \gptfivefour show strong familiarity with the mathematical literature and are often able to identify specialized results in published work. However, they struggle to locate more obscure sources, such as private lecture notes.

\vspace{-1mm}
\paragraph{Use of specialized tools} When confronted with complex computations, models frequently employ tools like SageMath. However, more specialized packages that are accessible through the bash tool pose challenges, with models often producing syntactically invalid code. After repeated failures, they sometimes revert to more common libraries, e.g., by using a manual Python re-implementation.

\vspace{-1mm}
\paragraph{Mistakes are often hidden} Models are typically quite economical with their mistakes, adding just a single simplifying assumption or incorrect claim. This one mistake often makes the problem significantly easier but leads to incorrect conclusions. Importantly, they are usually presented with confidence and framed rhetorically, for example, by stating that a ``well-known result'' implies a key step.  Sometimes, different models even independently converge on the \emph{same} shortcut, leading to parallel arguments that can create a false sense of consensus for the user. Although reasoning traces are often not accessible, we did not find evidence of \emph{deliberate deception} where models were aware of their own mistakes and presented the flawed argument nonetheless.

\vspace{-1mm}
\paragraph{Models rarely abstain} Models rarely abstain from claiming a solution to the presented IMProofBench questions. Even on the extremely challenging open problems in the benchmark, models almost always make an attempt at a definite answer. This happens despite user preferences strongly favoring an abstention over a mistaken but convincing proof.

\vspace{-1mm}
\paragraph{User testimonials} For many contributors, this benchmark was their first hands-on experience with state-of-the-art LLMs in an agentic setup. Participants at outreach events expressed surprise at the level of performance ({\it "Quite impressive, especially the case of degree 3 where one has to argue a little bit..."}). During grading, we found that some models applied new approaches to known problems, surprising the expert graders ({\it "I was not familiar with the correct solution from the models, even though it is relatively fundamental."}). Although only one open problem was solved, some other attempts also received positive feedback ({\it "Still I am amazed by the quality of the one-shot answers."}).

\vspace{-3mm}
\section{Limitations}\label{sec:limitations}

The main limitation of IMProofBench is its current scale, with only $\numberacceptedquestions$ questions included so far. However, we are continuously expanding the benchmark, with an additional $\numberquestionsdraft$ problems at an advanced draft stage and $\numberquestionsunderreview$ problems in the final stages of review. Even at this point, our analysis already provides detailed and valuable insights into the potential of LLMs for research-level mathematics, and these findings will become even more compelling as the benchmark develops further. Much smaller-scale evaluations of proof-based problems, such as those conducted on the USAMO and IMO 2025 \citep{usamo,matharena}, have already produced meaningful conclusions, which underscores the value of such efforts even when the number of problems is small.

\vspace{-2mm}
\section{Conclusion}\label{sec:conclusion}
In this paper, we introduced IMProofBench, a benchmark designed to evaluate research-level proof-writing capabilities in LLMs. Unlike prior datasets that focus primarily on final answers, IMProofBench evaluates whether models can produce logically sound arguments that meet the standards of mathematical research. Each problem is authored and peer-reviewed by professional mathematicians, and evaluation takes place in an agentic framework that mirrors a real research environment. Our experiments with state-of-the-art LLMs show that models can already solve a meaningful subset of research-level problems, with \gptfivefour achieving complete solutions on an estimated $49\%$ of tasks. These findings highlight that while current models remain imperfect and prone to errors, they are already capable of providing valuable support to working mathematicians for some problems.

\section*{Impact Statement}
The goals and impact of this benchmark are threefold. First, it aims to more accurately evaluate model performance on advanced, research-level mathematical problems within an evaluation framework that better reflects real-world usage. Second, we envision the benchmark as a tool for monitoring and studying how mathematicians can effectively leverage LLMs in their own research workflows, including identifying scenarios in which these models provide meaningful assistance on practical tasks. Third, it enables fine-grained analysis of model strengths and failure modes in advanced mathematical reasoning, helping guide future model development and training strategies.

\ificmlshowauthors
    \section*{Acknowledgments}

\paragraph{Acknowledgments} We would like to thank Honglu Fan for his help and support throughout the project, including advice and feedback at the early planning stages. We are also deeply grateful to the IT Support group of the D-MATH at ETH, and in particular to Michele Marcionelli, who provided competent, patient and prompt support in setting up our \href{https://improofbench.math.ethz.ch/}{benchmarking website}.

We also thank the organizers of the 2025 \href{https://people.math.ethz.ch/~rahul/ChAGS25.html}{Helvetic Algebraic Geometry Seminar}, Oberwolfach Workshop \href{https://publications.mfo.de/bitstream/handle/mfo/4289/OWR_2025_28.pdf#page=48}{"Recent Trends in Algebraic Geometry"} and the ITS workshop \href{https://eth-its.ethz.ch/activities/CAG.html}{"Computations in Algebraic Geometry: Complex, Real, and Tropical"} for providing outreach platforms to advertise the benchmark in the form of satellite events and during poster sessions. 

\paragraph{Financial acknowledgments (Project)} We acknowledge that our project received support in the form of free credits for both the xAI and the Gemini APIs, for which we thank the teams at the respective companies. The project also received an unrestricted research gift from Google to ETH Zurich (USD 30,000, November 2025).

\paragraph{Financial acknowledgments (Contributors)} Various contributors were supported by grants:
\begin{itemize}
\item David Holmes is supported by ERC grant EAGL.
\item Baran Hashemi is supported by the Excellence Cluster ORIGINS, funded by the Deutsche Forschungsgemeinschaft (DFG, German Research Foundation) under Germany's Excellence Strategy-EXC2094-390783311.
\item Chiara Meroni is supported by Dr.\ Max Rössler, the Walter Haefner Foundation, and the ETH Zürich Foundation.
\item Johannes Schmitt was supported by SNSF grant 200020-219369 and 200020-1000912 and SwissMAP.
\item Charles Vial is funded by the Deutsche Forschungsgemeinschaft (DFG, German Research Foundation) -- Project-ID 491392403 -- TRR 358.
\item Gergely Bérczi was supported by DFF Grant 40296.
\item Yannik Schuler is supported by a Walter-Benjamin fellowship of the DFG, project number 576663726, and SNSF grant 200020-219369.  %
\item Martijn Kool is supported by ERC Consolidator Grant FourSurf 101087365.  %
\item Pim Spelier is supported by ERC Consolidator Grant FourSurf 101087365.
\item David Muñoz-Lahoz is supported by an FPI-UAM 2023 contract funded by Universidad Autónoma de Madrid.
\item Ignacio Barros was supported by the Research Foundation -- Flanders (FWO) within the framework of the Odysseus program, project number G0D9323N, and by the Deutsche Forschungsgemeinschaft (DFG, German Research Foundation) -- Project-ID 491392403 -- TRR 358.
\item Aleksandar Mijatović is supported by EPSRC grants EP/V009478/1 and EP/W006227/1.
\item João Camarneiro's work was supported by the UKRI Centre for Doctoral Training in Algebra, Geometry and Quantum Fields (AGQ), Grant Number EP/Y035232/1.
\item Sergei Monavari was supported by the grant HORIZON-MSCA-2024-PF-01 Project 101203281-MoSSGIn.
\item Ronald van Luijk is supported by NWO XL grant OCENW.XL21.XL21.011.
\item Pieter Belmans ~was partially supported by NWO (\href{https://doi.org/10.61686/RZKLF82806}{\texttt{doi:10.61686/RZKLF82806}}).
\end{itemize}

\else
    \ifarxiv
        
    \fi
\fi

\message{^^JLASTBODYPAGE \thepage^^J}

\bibliographystyle{icml2026}
\bibliography{references}

\message{^^JLASTREFERENCESPAGE \thepage^^J}

\appendix
\onecolumn
\crefalias{section}{appendix}
\crefalias{subsection}{appendix}
\crefalias{subsubsection}{appendix}
\section{Benchmark Composition and Additional Evaluation Results}\label{app:details}

\subsection{Extended Results}

\begin{figure}[htbp]
    \centering

        \includegraphics[width=\linewidth]{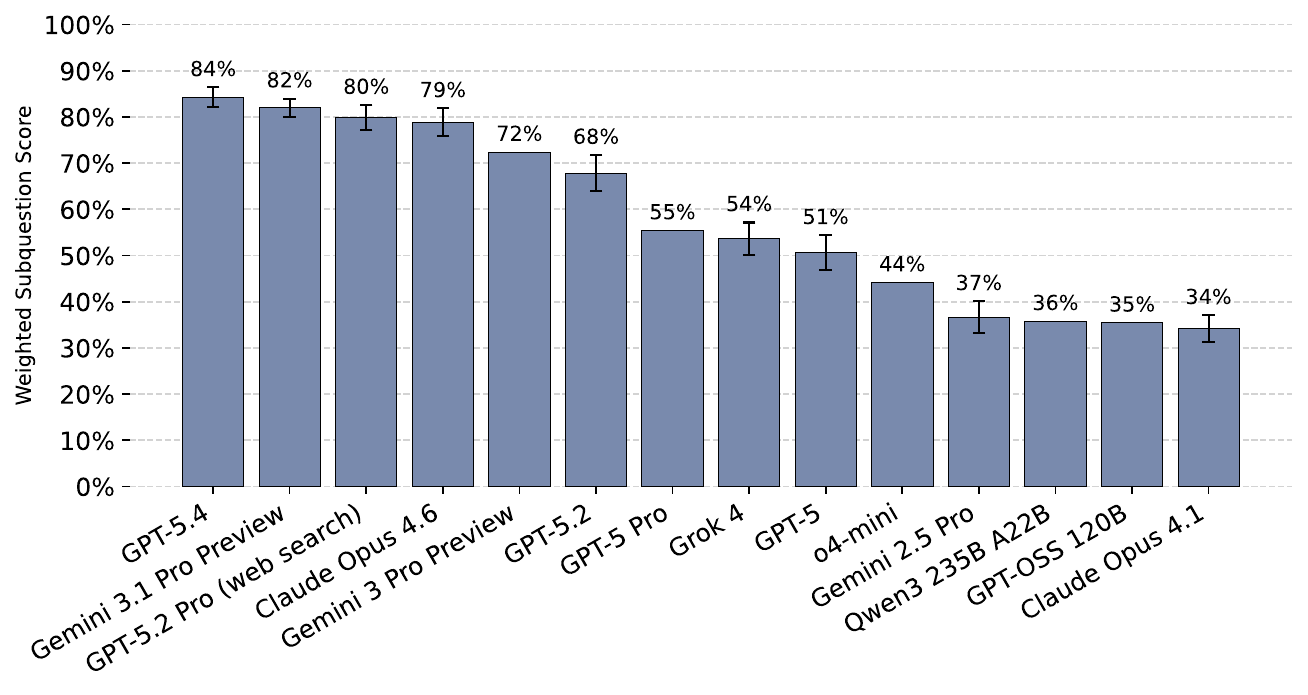}
    \caption{Average percentage of points for subquestion evaluation. Here, performance on any individual question is weighted by the point rewards determined by the problem author. $1\sigma$ confidence intervals are reported for models that were executed at least twice on the benchmark.}
    \label{fig:subquestions_weighted}
\end{figure}

\begin{figure}[htbp]
    \centering

        \includegraphics[width=\linewidth]{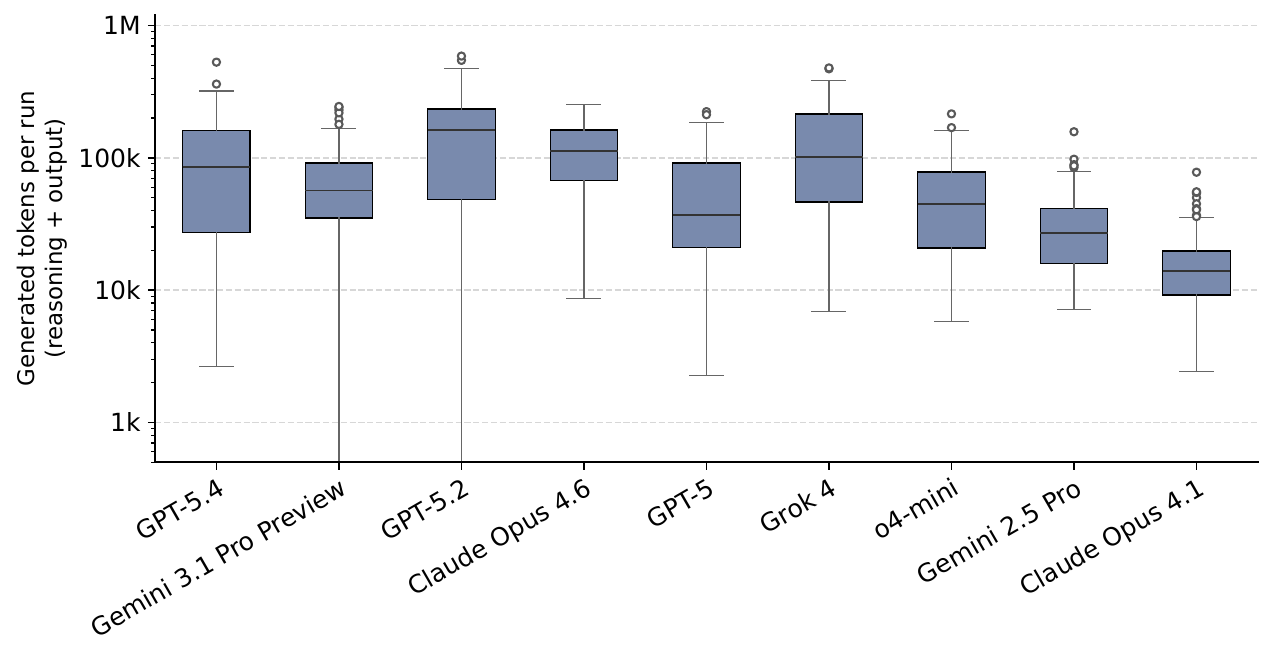}
    \caption{Token usage distribution for problem evaluation (main question and subquestions) for the models evaluated in our agentic framework, with reasoning and output tokens combined.}
    \label{fig:tokenuse_logarithmic}
\end{figure}

\begin{figure}[htbp]
    \centering

        \includegraphics[width=\linewidth]{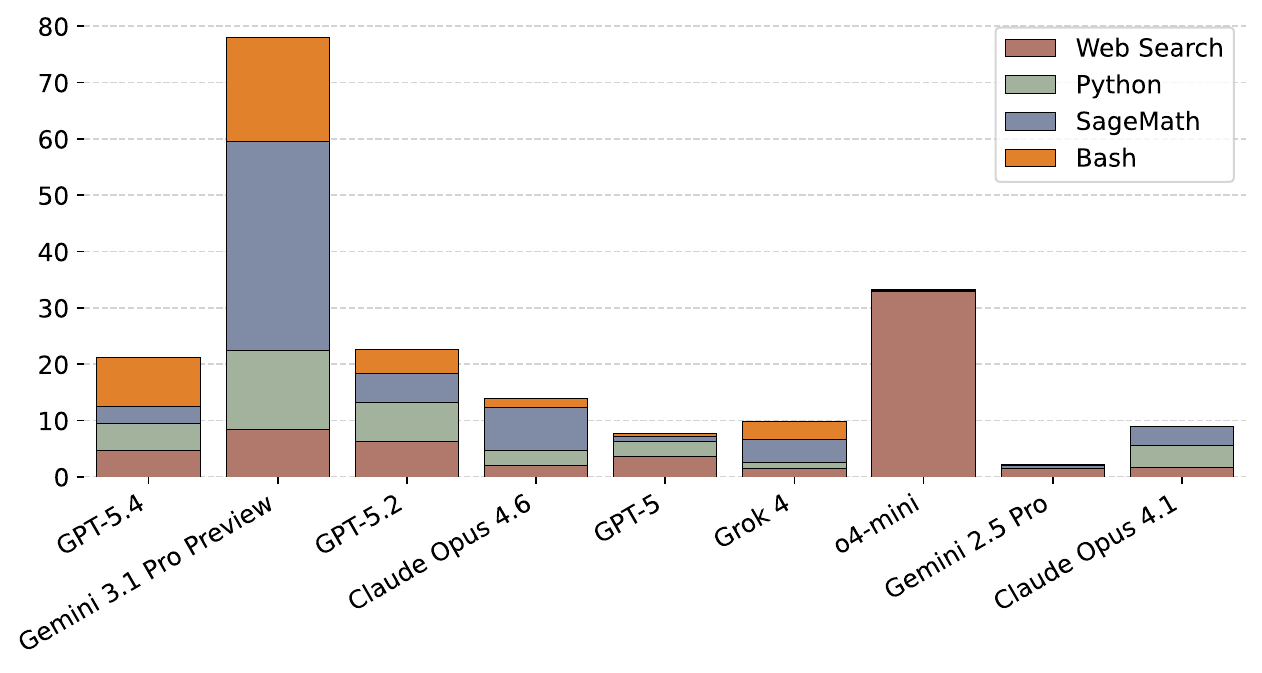}
    \caption{Average tool usage for the models evaluated in our agentic framework.}
    \label{fig:toolusage_all}
\end{figure}

\paragraph{Performance on final-answer subquestions}
In \cref{fig:subquestions_weighted}, we present the average scores obtained by $\numberevaluatedmodels$ evaluated models on the final-answer subquestions, using the author-appointed weights that reflect importance or difficulty. As shown in the figure, \gptfivefour obtains the best performance with a score of $84\%$.

\paragraph{Token usage}
In \cref{fig:tokenuse_logarithmic}, we show the distribution of generated tokens (reasoning and output combined) across the evaluated questions. With respect to token limits, which allow 300k tokens for the main question and 100k tokens for each subquestion, models almost always remain below these thresholds, except for \gptfivetwo, which exceeded 300k tokens on the main question in $9\%$ of runs.

\paragraph{Tool usage}
In \cref{fig:toolusage_all}, we show the average tool usage across models. The patterns differ substantially. \ofour makes around $33$ tool calls per problem, relying more heavily on the web search tool than any other model. Further, only more recent models have started to use the bash tool heavily.

\subsection{Topics in IMProofBench}
\begin{figure}[htbp]
    \centering

    \includegraphics[width=\linewidth]{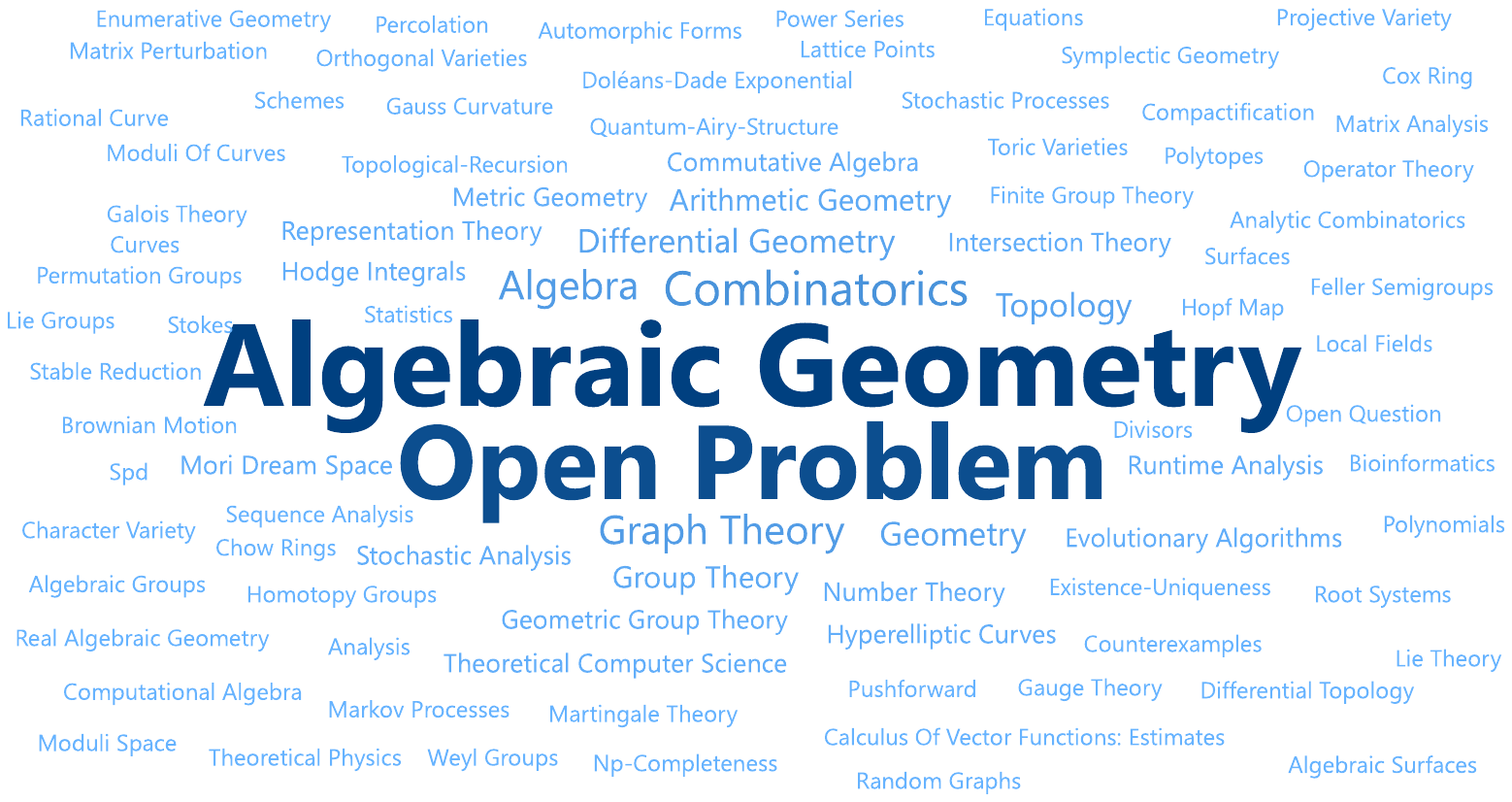}
    \vspace{-24mm}
    \caption{Word cloud of tags assigned to IMProofBench problems.}
    \label{fig:wordcloud}
\end{figure}

In \cref{fig:wordcloud}, we display the distribution of problem tags in IMProofBench. The topic of "Algebraic Geometry" currently dominates, reflecting the research focus of the benchmark organizers. These organizers both contributed problems themselves and solicited input primarily from colleagues in their own academic networks. Future development of the benchmark will aim to broaden its coverage to include a wider range of topics in pure and applied mathematics, as outlined in \cref{app:development}.

\subsection{Analysis of human grading reliability} \label{app:grading_reliability}
To assess inter-rater reliability of our human grading scheme, we conducted an analysis during which $\agreementquestions$ questions were independently graded by multiple evaluators, yielding $\agreementcomparisons$ pairwise judgment comparisons.

For the overall progress grade (scored 0-3), graders achieved exact agreement in $\progressexactagreement\%$ of cases and agreed within one point in $\progresswithinoneagreement\%$ of cases, suggesting that the coarse-grained progress assessment is moderately reliable (Cohen's $\kappa = 0.40$). At the question level, $\agreementhigh\%$ of questions showed high overall agreement ($\geq 80\%$), $\agreementmoderate\%$ showed moderate agreement ($60-79\%$), and only $\agreementlow\%$ showed low agreement ($<60\%$). Manual inspection of these discrepancies indicates that the secondary grader was sometimes not aware that a result cited by the LLM actually exists and applies, and thereby graded the solution as incorrect. Since the actual primary graders are better domain experts for their own question, we expect this issue is much less likely to occur for our actual graders.

Agreement varied considerably across the eight binary grading categories, as shown in \cref{Table:categoryagreement}. Objective categories such as ``Correct End Result'' showed high agreement ($\agreementcorrectendresult\%$), while more subjective error classifications like ``Incorrect Logic'' and assessments of ``Mathematical Insight'' proved harder to pin down ($\agreementincorrectlogic\%$ and $\agreementmathematicalinsight\%$ respectively). This suggests that while graders largely agree on whether a solution succeeds and makes meaningful progress, diagnosing the precise nature of errors in failed attempts involves more subjective judgment.

\begin{table}[ht]
\centering
\caption{Inter-rater agreement by grading category, based on $\agreementcomparisons$ pairwise comparisons across $\agreementquestions$ questions, excluding comparisons where either grader selected ``Not Sure'' or ``N/A''.}
\label{Table:categoryagreement}
\setlength{\tabcolsep}{10pt}
\renewcommand{\arraystretch}{1.3}
\begin{tabular}{lc}
\toprule
Category & Agreement \\
\midrule
Correct End Result & $\agreementcorrectendresult\%$ \\
Problem Understanding & $\agreementproblemunderstanding\%$ \\
Useful Progress & $\agreementusefulprogress\%$ \\
Hallucinated Facts & $\agreementhallucinatedfacts\%$ \\
Conceptual Error & $\agreementconceptualerror\%$ \\
Calculation Error & $\agreementcalculationerror\%$ \\
Mathematical Insight & $\agreementmathematicalinsight\%$ \\
Incorrect Logic & $\agreementincorrectlogic\%$ \\
\bottomrule
\end{tabular}
\end{table}

\section{Human Interface and Instructions}\label{app:interface}
In this appendix, we discuss how contributors and benchmark administrators interact with IMProofBench, including the instructions and interface for different steps of the submission process (question generation, review, and grading). In \cref{app:interface:submission}, we give a brief overview of the main pages on the web interface. Then, in \cref{app:question_creation}, we provide details on how questions are created and edited. In \cref{app:review_process}, we explain the review process. Finally, in \cref{app:grading_interface}, we discuss the grading interface.

\subsection{Submission Website} \label{app:interface:submission}
\begin{figure}[t]
    \centering
        \includegraphics[width=0.85\linewidth]{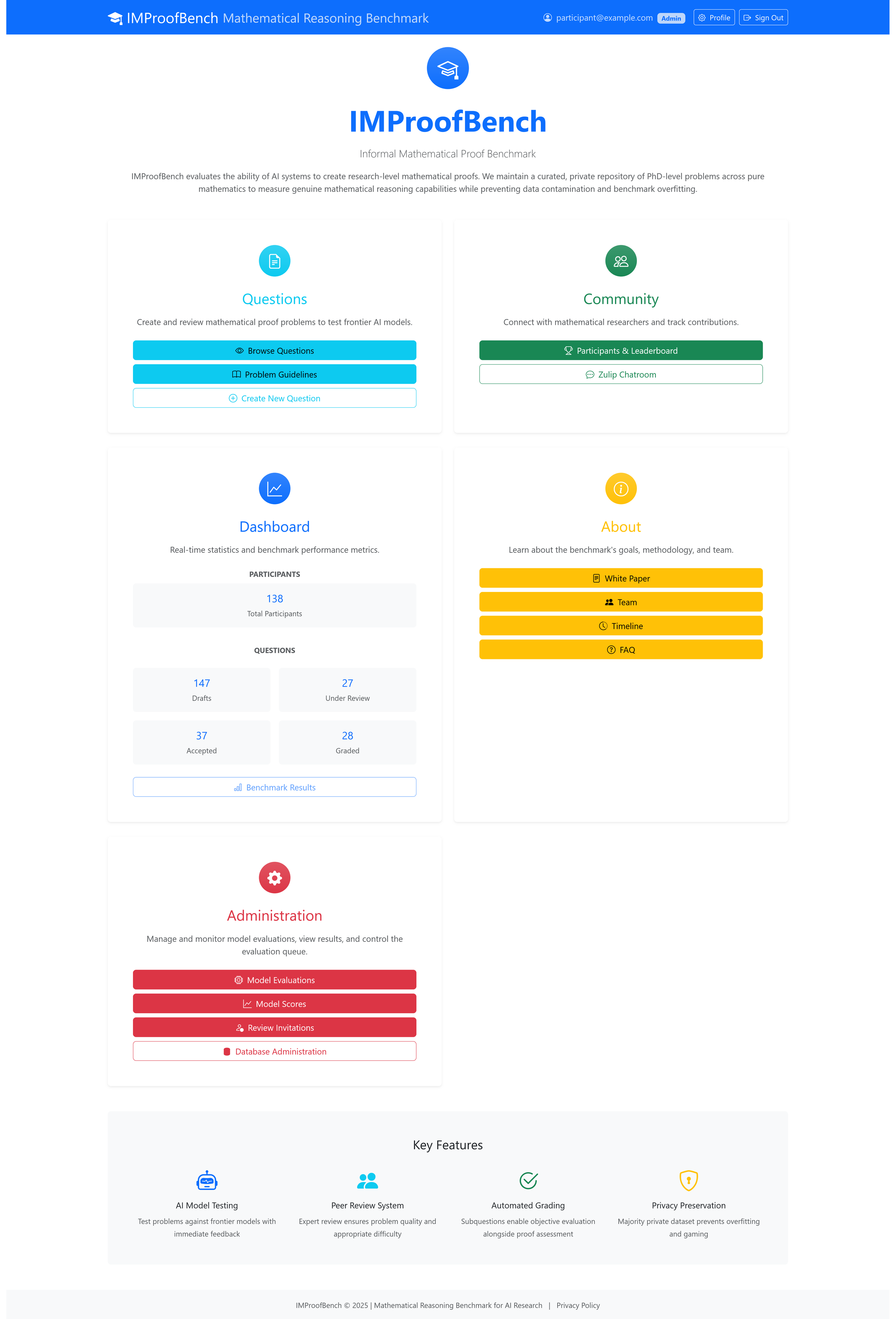}
    \caption{Landing and overview page of IMProofBench website.}
    \label{fig:screen:landing}
\end{figure}

Contributors submit problems via a secure website designed for submitting and reviewing questions, and grading AI answers (see \cref{fig:screen:landing}). Features include:
\begin{itemize}[left=1em,itemsep=0.1em]
    \item \textbf{User accounts and permissions}: Contributors can create an account tied to a (verified) email, which allows them to author questions and use website features like the free AI solution previews for these questions. Benchmark administrators have additional access to manage model evaluations, review requests, and access a live view of benchmark results.
    \item \textbf{Community features}: The website shows a list of contributors (ordered by the number of accepted questions or similar parameters) to encourage active participation, and links to a project Zulip with further news and an opportunity to provide feedback.
    \item \textbf{Benchmark dashboard}: Total numbers of contributors and questions in different stages of the submission process are displayed to show project progress. An overview page with both live results and archived snapshots of the benchmark state will be added in the future.
    \item \textbf{About the project}: Information about the IMProofBench is provided. This information contains the initial whitepaper, an overview of core team members, a timeline of planned steps, and a page with frequently asked questions. A privacy policy detailing our handling of user data is linked in the footer of the page.
\end{itemize}

\subsection{Question Creation and Editing} \label{app:question_creation}

\begin{figure}[t]
    \centering
        \includegraphics[width=0.75\linewidth]{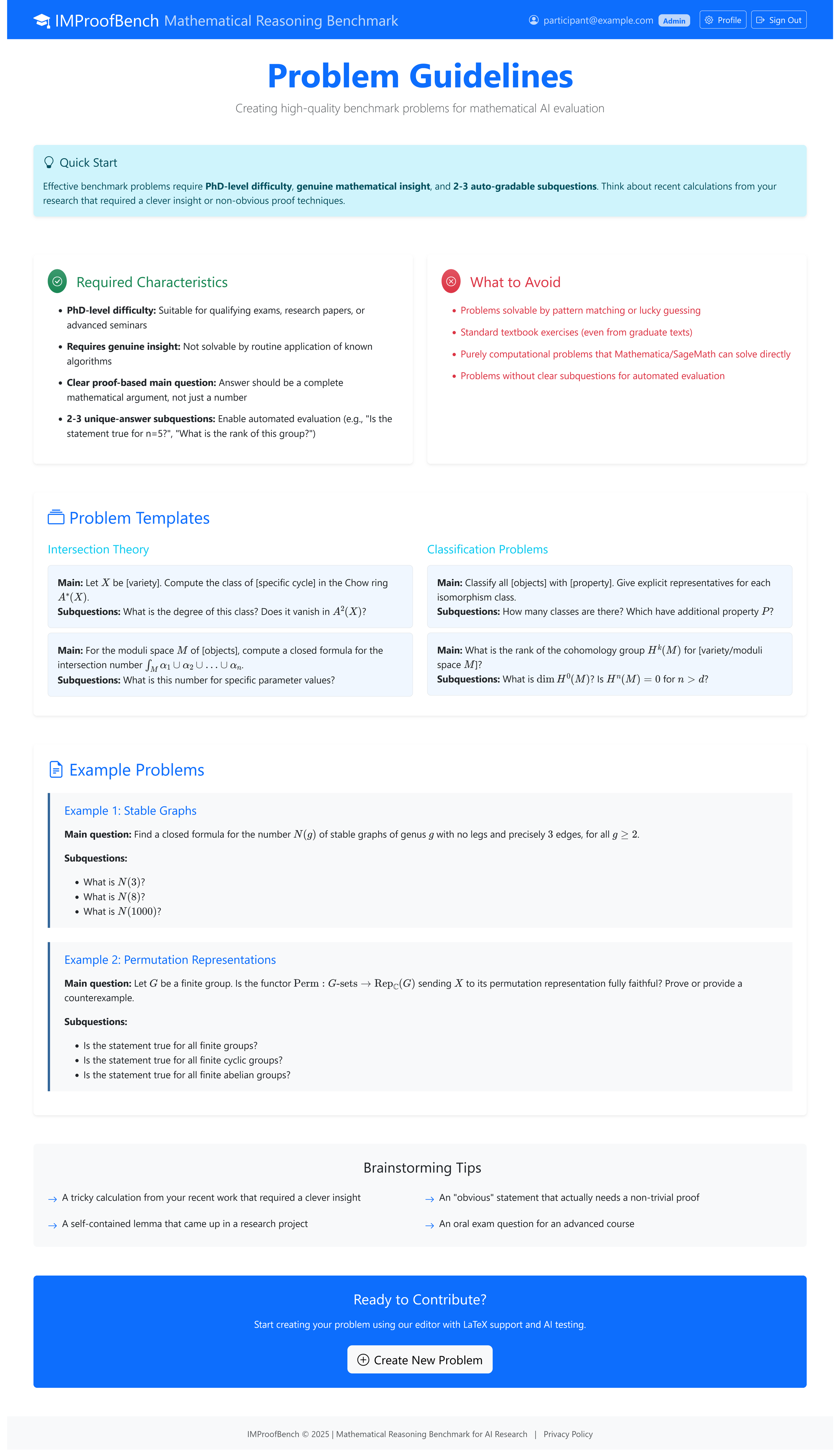}
    \caption{Guidelines for authoring benchmark problems.}
    \label{fig:screen:problemguidelines}
\end{figure}

Benchmark problems are created through a structured interface that guides contributors through the submission requirements. The system provides comprehensive guidelines (see Figure~\ref{fig:screen:problemguidelines}) emphasizing the key characteristics of suitable benchmark problems.

\paragraph{Problem guidelines} Effective benchmark problems must meet several criteria:
\begin{itemize}[left=1em,itemsep=0.1em]
    \item \textbf{PhD-level difficulty:} Problems should be suitable for oral exams of graduate courses, research papers, or advanced seminars, representing mathematics close to or at research-level.
    \item \textbf{Genuine mathematical insight:} Solutions must require non-routine approaches that cannot be solved through pattern matching or standard algorithm application.
    \item \textbf{Clear proof-based main question:} The primary answer should consist of a complete mathematical argument rather than merely a numerical result.
    \item \textbf{Auto-gradable subquestions:} Each problem ideally includes 2--3 subquestions with unique answers (e.g., ``Is the statement true for $n=5$?'' or ``What is the rank of this group?''), enabling automated evaluation.
\end{itemize}

Contributors should avoid problems solvable by lucky guessing, standard textbook exercises (even from graduate texts), or purely computational problems that mathematical software can solve directly.

\begin{figure}[t]
    \centering
        \includegraphics[height=0.9\textheight]{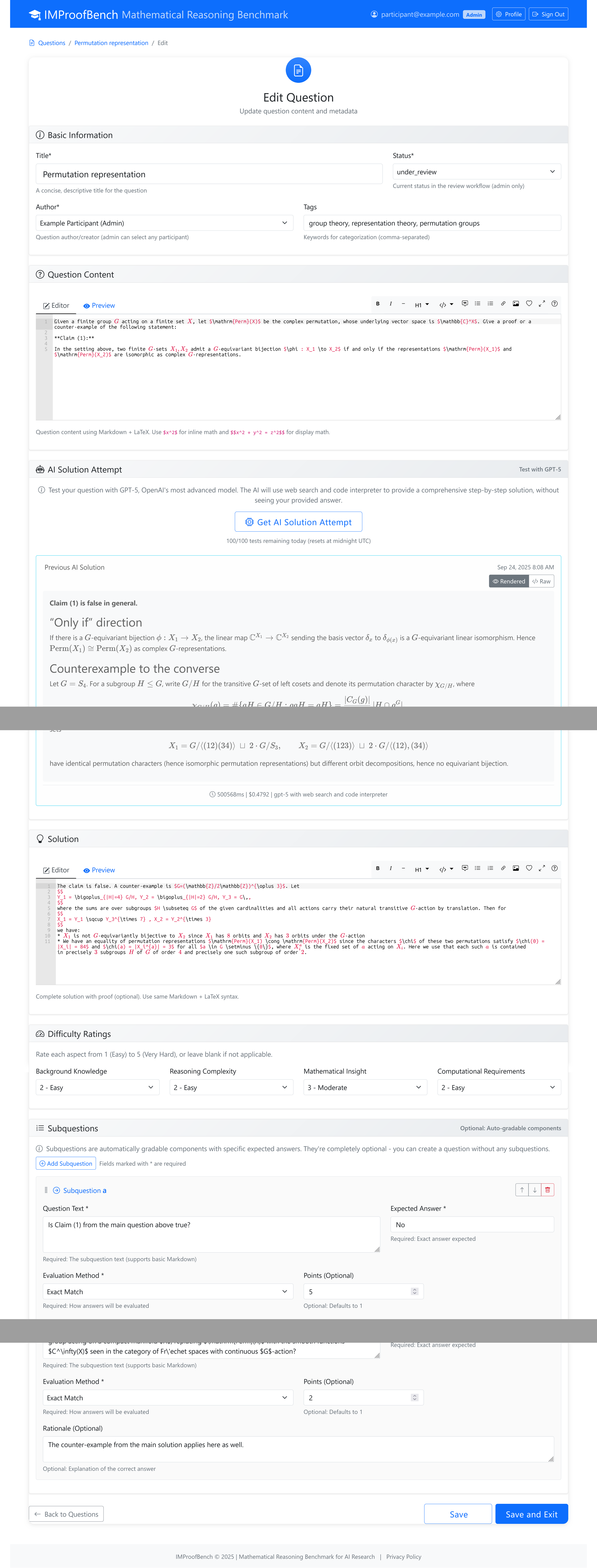}
    \caption{Window for editing questions, solutions, and their associated subquestions; via the blue button, the user can request up to 20 free AI solution previews per day to check the suitability of the question.}
    \label{fig:screen:questionedit}
\end{figure}

\paragraph{Question editing interface} The question creation and editing window (see Figure~\ref{fig:screen:questionedit}) provides a comprehensive authoring environment with the following components:

\begin{itemize}[left=1em,itemsep=0.1em]
    \item \textbf{Main question editor:} A text area supporting Markdown with LaTeX mathematics, featuring a live preview pane that renders the formatted content in real-time. Contributors can use standard LaTeX delimiters (\texttt{\$...\$} for inline and \texttt{\$\$...\$\$} for display mathematics).
    
    \item \textbf{Problem metadata:} A tags field allows contributors to categorize problems by area (e.g., ``group theory'', ``representation theory'', or ``permutation groups'') and special characteristics (e.g., ``open problem'' for questions where the author seeks but does not know the answer).
    
    \item \textbf{AI solution preview:} Contributors can test their questions against a frontier AI model (currently \gptfivefive with high reasoning effort) using up to 20 free attempts per day. This feature helps authors evaluate whether their problem has appropriate difficulty and clarity.
    
    \item \textbf{Sample solution:} A dedicated editor for the complete solution, which serves as the reference for reviewers and graders. The solution should demonstrate the expected level of rigor and detail to allow expert review to verify correctness and serve as a reference for grading model answers.
    
    \item \textbf{Subquestions management:} A dynamic form system for adding multiple subquestions, where each subquestion consists of:
    \begin{itemize}
        \item Question text (supporting Markdown and LaTeX)
        \item Expected answer field for the unique answer
        \item Evaluation method selector (e.g., exact match)
        \item Optional points value (defaulting to 1) for weighting subquestions by difficulty or importance
        \item Rationale field for explaining the correct answer
    \end{itemize}
\end{itemize}

\begin{figure}[t]
    \centering
        \includegraphics[width=0.7\linewidth]{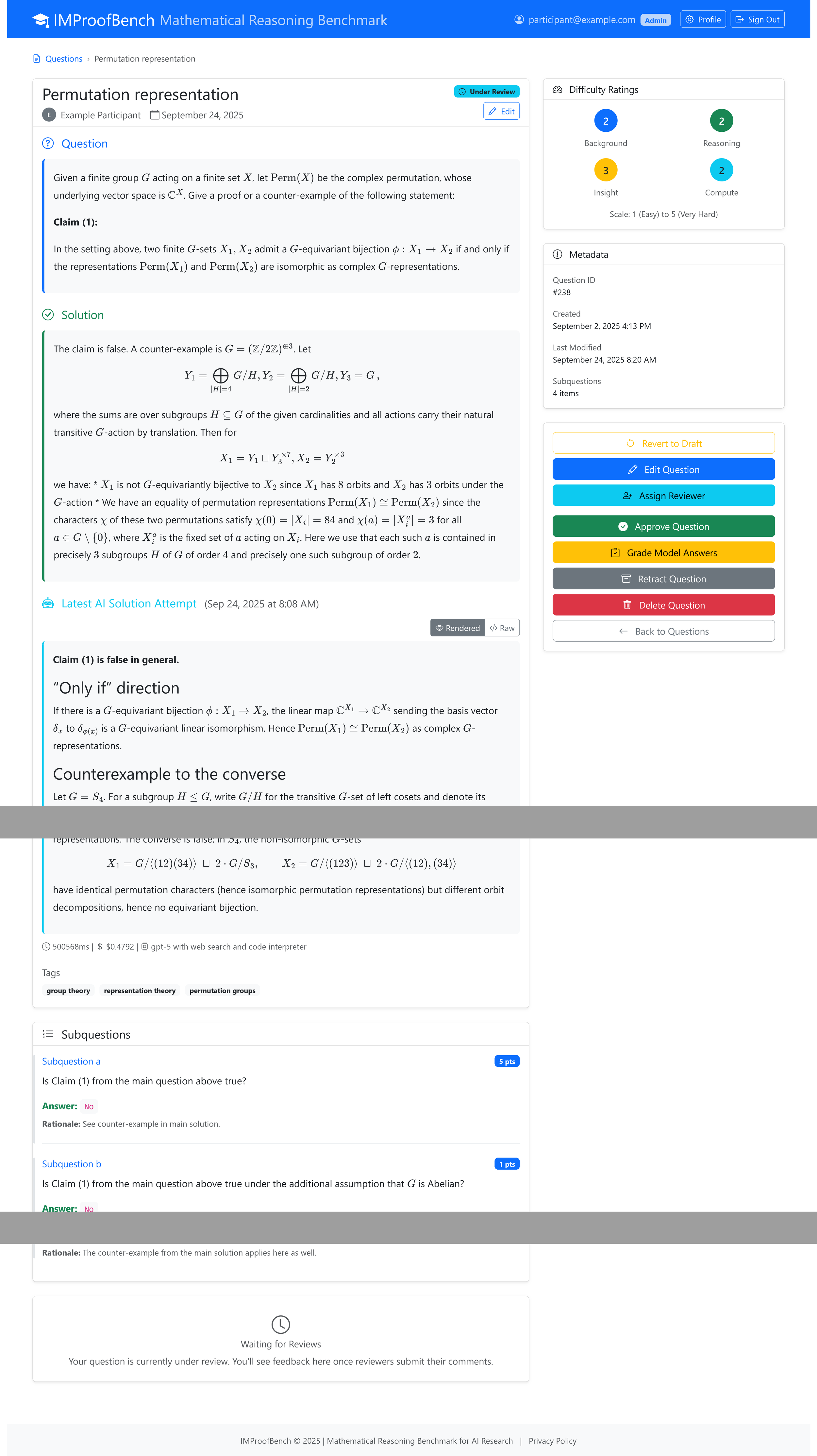}
    \caption{Overview page of question data (with main question, sample solution, AI answer preview, and subquestions).}
    \label{fig:screen:questiondetail}
\end{figure}

\paragraph{Question detail view} Once submitted, questions are displayed in a detail view (see Figure~\ref{fig:screen:questiondetail}) that presents all components in their rendered form. This view shows:
\begin{itemize}[left=1em,itemsep=0.1em]
    \item The question status in the submission pipeline (Draft $\rightarrow$ Under Review $\rightarrow$ Approved $\rightarrow$ Active)
    \item Rendered the main question and sample solution with properly formatted mathematics
    \item List of subquestions with their expected answers
    \item AI solution attempt preview when available
    \item Review comments from expert reviewers (when in review stage)
    \item Response interface allowing authors to address reviewer feedback and revise their submission
\end{itemize}

The detail view serves as the central hub for tracking a question's progress through the review process and facilitating communication between authors and reviewers.

\subsection{Review Process and Instructions} \label{app:review_process}

\begin{figure}[t]
    \centering
        \includegraphics[width=0.75\linewidth]{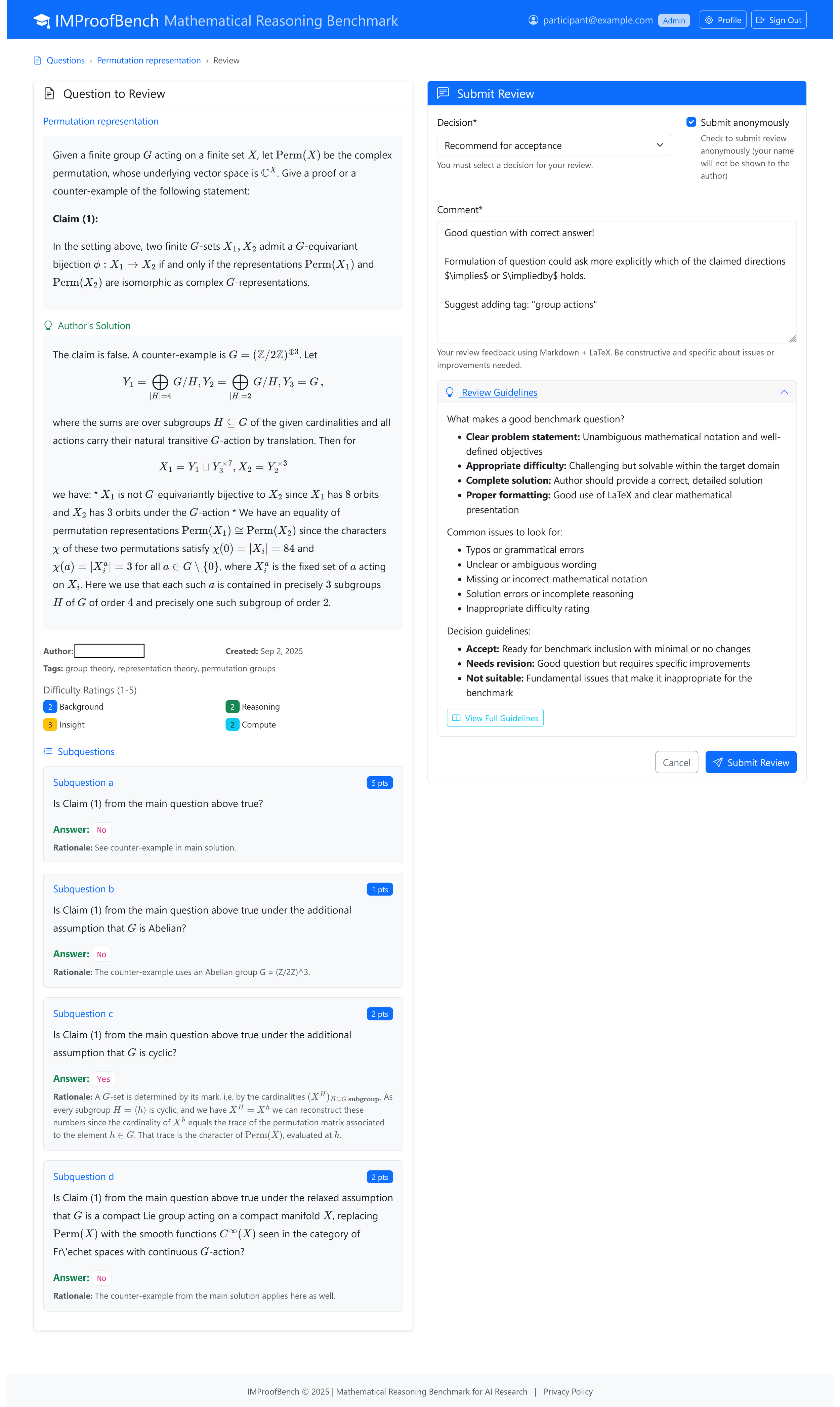}
    \caption{Question review window showing text box for feedback and review instruction summary.}
    \label{fig:screen:review}
\end{figure}

\begin{figure}[t]
    \centering
        \includegraphics[width=0.85\linewidth]{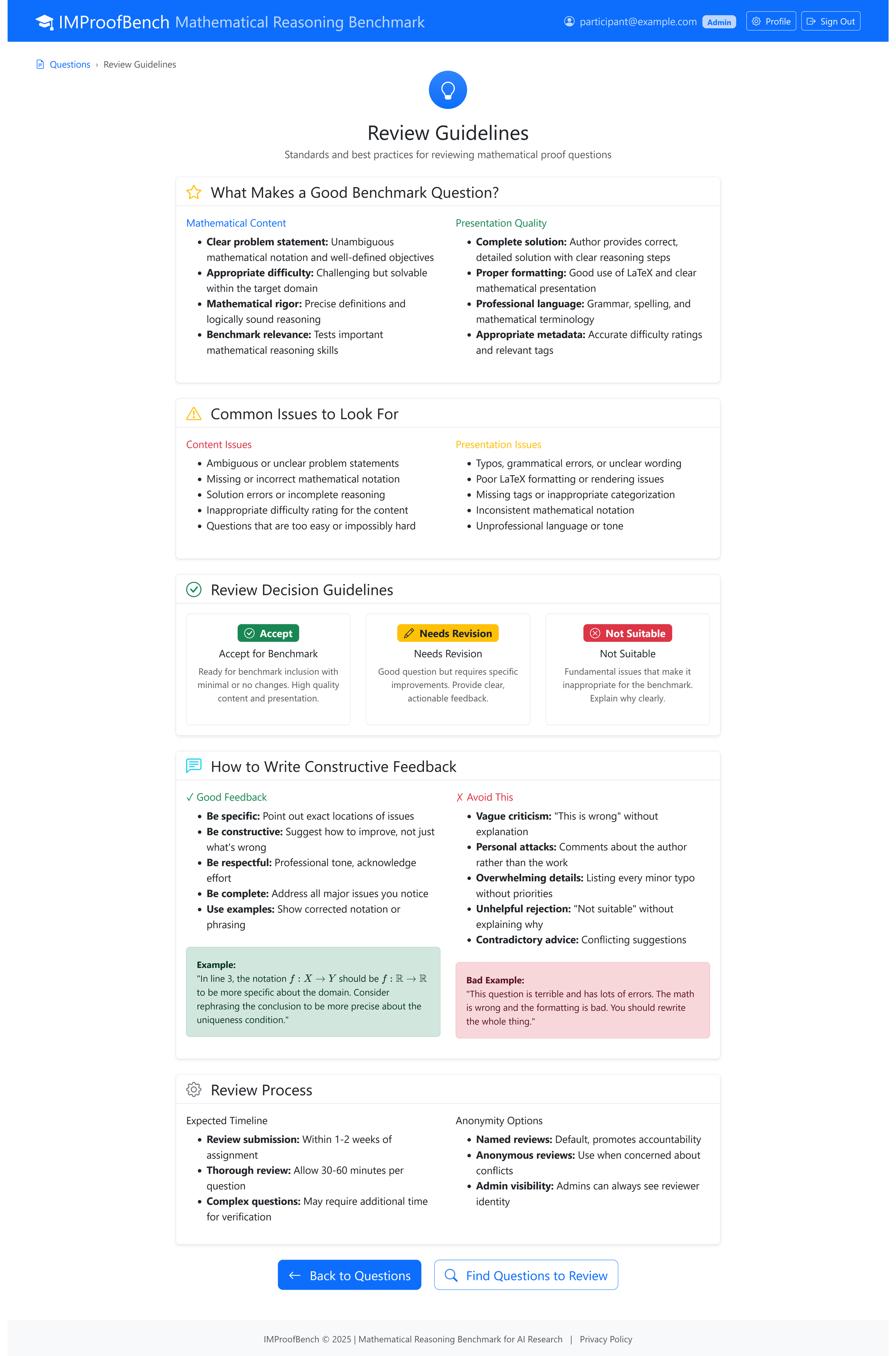}
    \caption{Detailed explainer of review instructions and process.}
    \label{fig:screen:reviewinstructions}
\end{figure}

Each question is reviewed by at least one expert before being included in the benchmark. These experts are invited to submit
a review via email. An example of such
an email is included below.
\begin{prompt}{Reviewer invitation email}
Dear [invited_user],

My name is [inviting_user] and I am part of a small team of mathematicians studying the question of how good today's AI models are at solving research-level math questions. As part of this IMProofBench project, we are building a collection of challenging mathematical problems to use for testing the AI performance. 

We would like to ask for your help in verifying the mathematical correctness of one such question. If you are interested to learn more about the project, further information is available at [LINK]

The following question was submitted for inclusion in the IMProofBench dataset:

Title: Permutation representation
Author: Example Participant

Would you be willing to review this question and:
- Verify that the phrasing is well-defined and unambiguous
- Confirm the provided solution is mathematically correct
- Make any suggestions for improvements (e.g., additional unique-answer subquestions)

We estimate that for most problems, this should take between 10 and 30 minutes.

You can view the full submitted problem and write a review at:
[ACCEPT_URL]

There, you will also have the option to decline this review request after viewing the question.
Alternatively, you can decline immediately by clicking:
[DECLINE_URL]

If you provide a review, the question's author will be notified and have the chance to revise the question and compose a response. After seeing the response, you have the option to submit a further review or recommend the question for acceptance in the benchmark.

Thank you for considering this request!

Best regards,
[inviting_user]

Note: To track your review and allow you to see the author's replies, accepting the review request will create a user account for you on our website. You can optionally set a password after submitting your review to log back in and e.g., contribute a question to the benchmark yourself.
\end{prompt}
When the reviewer accepts the review invitation by clicking on the link, they are forwarded to a webpage displaying the problem to be reviewed, along with a form for review submission and further information (see Figure \ref{fig:screen:review}). The reviewer may also view the full review guidelines displayed in Figure \ref{fig:screen:reviewinstructions}. The review consists of a short comment by the reviewer indicating improvements and/or mistakes in the question statement. Before submitting the review, the reviewer decides on a recommended action among the following: ``Recommended for acceptance'', ``Needs revision'' and ``Not suitable''.
The site admins are notified when a review
is complete and can take action accordingly.
If the reviewer selects ``Not suitable'', the question is automatically reset to the ``draft'' status. Independent of the outcome, the author is permitted to submit an answer to the reviewer's comments and change the question if necessary. The reviewer may then either submit a new review taking into account the changes, or a new reviewer may be invited.

\subsection{Grading Interfaces} \label{app:grading_interface}

The grading system provides a structured interface for human evaluation of model-generated proofs through a dedicated web page.

\begin{figure}[t]
    \centering
        \includegraphics[width=\linewidth, trim = {46cm 0 46cm 0}, clip]{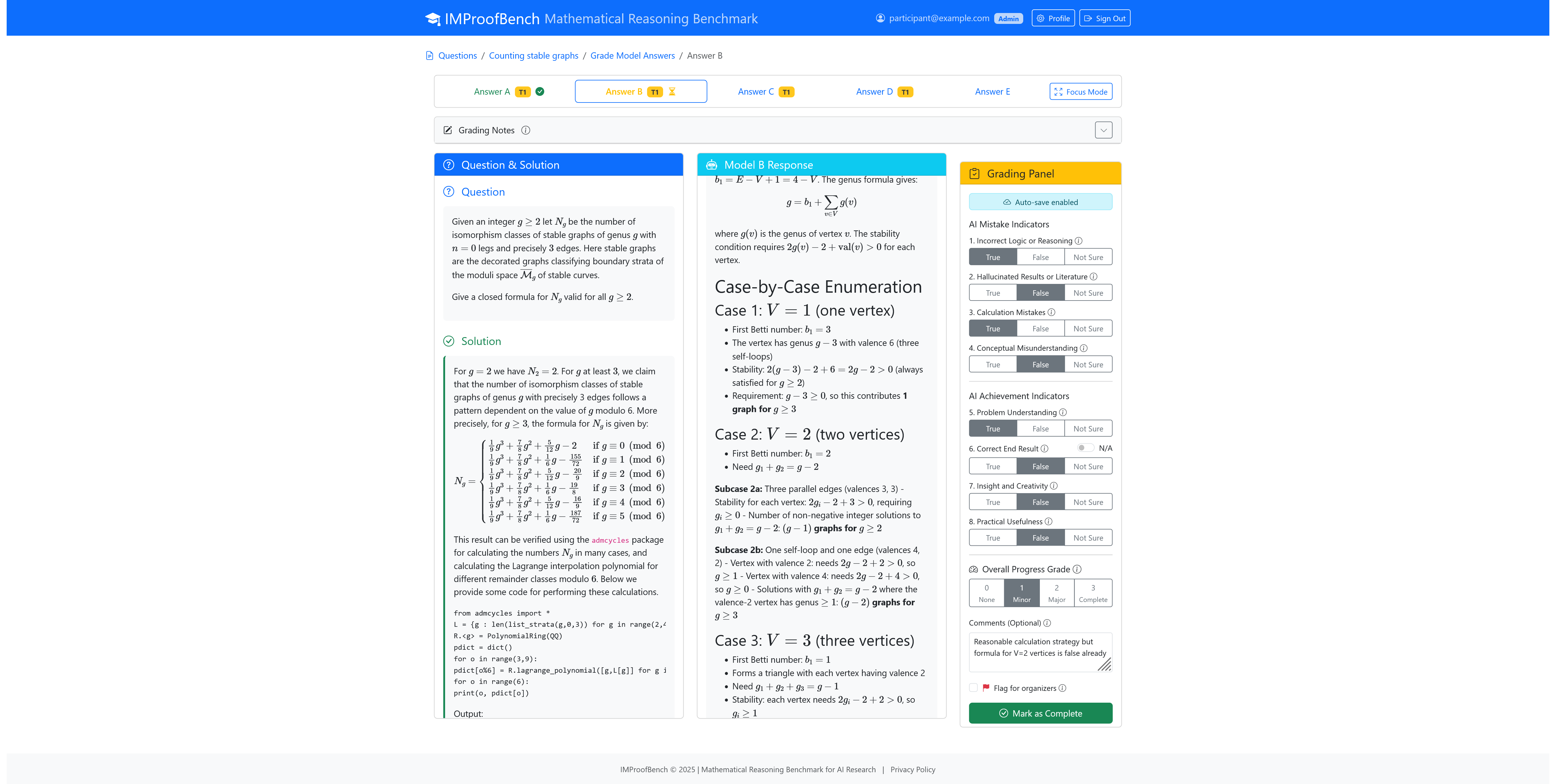}
    \caption{Grading form, displaying sample solution, model answer, and scoring form side by side. Model identities (A--E) at the top are randomized on starting the grading, and only revealed when grading is complete, to avoid bias.}
    \label{fig:screen:grading}
\end{figure}

\paragraph{Human grading interface} The main grading interface (see Figure~\ref{fig:screen:grading}) employs a three-column layout designed to facilitate easy access to relevant information and the feedback form:

\begin{itemize}[left=1em,itemsep=0.1em]
    \item \textbf{Left column:} Displays the question statement and sample solution for reference
    \item \textbf{Center column:} Shows the model's complete response with mathematical rendering
    \item \textbf{Right column:} Contains the interactive grading panel with scoring controls
\end{itemize}

To prevent bias, model identities are concealed behind randomized aliases (Answer A, B, C, etc.) that remain hidden until all answers for a question have been graded. The system maintains independent grading sessions for each evaluator, with aliases shuffled differently to ensure blind evaluation.

\paragraph{Grading categories} The scoring form consists of three main components providing multifaceted evaluation, with relevant information available via concise tooltips:

\textbf{AI mistake indicators:} Four binary categories identifying common failure modes:
\begin{enumerate}[left=1em,itemsep=0.1em]
    \item \textbf{Incorrect Logic:} Flawed logical steps or invalid reasoning
    \item \textbf{Hallucinated:} References to non-existent theorems, papers, or results
    \item \textbf{Calculation:} Arithmetic or algebraic errors
    \item \textbf{Conceptual:} Fundamental misunderstanding of mathematical concepts
\end{enumerate}

\textbf{AI achievement indicators} Four binary categories recognizing positive aspects:
\begin{enumerate}[left=1em,itemsep=0.1em]
    \setcounter{enumi}{4}
    \item \textbf{Understanding:} Correctly identifies what needs to be proven or calculated
    \item \textbf{Correct Result:} Arrives at the correct final answer (with N/A option for open-ended problems or when the correct answer is unknown)
    \item \textbf{Insight:} Shows creative problem-solving or novel approaches
    \item \textbf{Usefulness:} Solution would be helpful to someone learning this topic
\end{enumerate}

Each binary category offers three response options: ``True'', ``False'', or ``Not Sure'', allowing graders to indicate uncertainty when evaluation is ambiguous.

\textbf{Overall progress} A four-point scale (0--3) rating overall solution progress:
\begin{itemize}[left=1em,itemsep=0.1em]
    \item \textbf{0/3:} No progress toward solution
    \item \textbf{1/3:} Minor progress with limited advancement
    \item \textbf{2/3:} Major progress with substantial work completed
    \item \textbf{3/3:} Complete solution achieved
\end{itemize}

This overall progress score serves as the primary metric for model ranking and comparison.

\paragraph{Additional grading features} The interface includes several supporting elements to ensure grading consistency and quality:

\begin{itemize}[left=1em,itemsep=0.1em]
    \item \textbf{Grading notes:} A persistent text area where graders record their evaluation criteria and decision patterns across all answers (e.g., ``Matrix errors count as Calculation, Theory errors as Logic''). These notes help maintain consistency when grading multiple model responses and facilitate reproducibility in future grading sessions.
    
    \item \textbf{Comments field:} Answer-specific observations about edge cases or explanations for grading decisions.
    
    \item \textbf{Auto-save functionality:} Grading selections are automatically preserved with a 2-second debounce to prevent data loss.
    
    \item \textbf{Focus mode:} An optional distraction-free interface that maximizes screen space by hiding navigation elements and allowing collapsible panels, enabling graders to concentrate on detailed evaluation.
    
    \item \textbf{Flag for organizers:} Option to mark responses requiring special attention due to serious issues or technical problems.
\end{itemize}

The grading workflow supports iterative evaluation, allowing graders to mark answers as complete, incomplete, or given up (for responses that cannot be meaningfully evaluated). Once all model answers for a question are marked complete, the system reveals the true model identities, enabling post-hoc analysis of performance patterns.

\section{Sample Problem}
\label{app:sampleProb}
Below, we present an example of a problem from the benchmark and discuss model performance and solution strategies from our evaluation.
\paragraph{Background for reader (not included in benchmark question)}
A \emph{stable graph} is a connected graph $\widehat \Gamma$, multi-edges and loops allowed, together with a vertex-labeling by non-negative integers $(g_v)_{v \in V(\widehat \Gamma)}$ satisfying that each vertex $v$ with $g_v=0$ has valence at least $3$. These combinatorial objects appear in algebraic geometry in the study of moduli spaces of stable curves, see e.g. Section 2 of \cite{Schmittvanzelm}. The \emph{genus} of $\widehat \Gamma$ is defined as $g=b_1(\widehat \Gamma) + \sum_{v \in V(\widehat \Gamma)} g_v$, with $b_1$ the first Betti number (or cyclomatic number) of $\widehat \Gamma$.

\paragraph{Question}
Given an integer $g \geq 2$, let $N_g$ be the number of isomorphism classes of stable graphs of genus $g$ with precisely $3$ edges. Give a closed formula for $N_g$ valid for all $g \geq 2$.

\paragraph{Solution}
To compute $N_g$, we note that each stable graph $\widehat \Gamma$ has an undecorated underlying graph $\Gamma$, which is one of the $10$ connected multi-graphs with precisely $3$ edges. Then $N_g$ can be calculated by summing over those graphs $\Gamma$ and counting the number of assignments $g_v$ to the vertices of $\Gamma$, avoiding double-counting by taking into account symmetries of $\Gamma$.

The final answer is that for $g = 2$ we have $N_2 = 2$ and for $g \geq 3$, we have
$$N_g = \begin{cases}
\frac{1}{9}g^3 + \frac{7}{8}g^2 + \frac{5}{12}g - 2 & \text{if } g \equiv 0 \pmod{6} \\
\frac{1}{9}g^3 + \frac{7}{8}g^2 + \frac{1}{6}g - \frac{155}{72} & \text{if } g \equiv 1 \pmod{6} \\
\frac{1}{9}g^3 + \frac{7}{8}g^2 + \frac{5}{12}g - \frac{20}{9} & \text{if } g \equiv 2 \pmod{6} \\
\frac{1}{9}g^3 + \frac{7}{8}g^2 + \frac{1}{6}g - \frac{19}{8} & \text{if } g \equiv 3 \pmod{6} \\
\frac{1}{9}g^3 + \frac{7}{8}g^2 + \frac{5}{12}g - \frac{16}{9} & \text{if } g \equiv 4 \pmod{6} \\
\frac{1}{9}g^3 + \frac{7}{8}g^2 + \frac{1}{6}g - \frac{187}{72} & \text{if } g \equiv 5 \pmod{6}
\end{cases}$$

\paragraph{Subquestions}
What is $N_3$? (Answer: 9)
What is $N_8$? (Answer: 114)
What is $N_{10000}$? (Answer: 111198615276)

\paragraph{Model approaches and performance}
\begin{itemize}[left=1em,itemsep=0.1em]
    \item \gptfive instantly identifies the solution strategy in its first reasoning step, writing {\it "Essentially, I’m computing $N_g$ as a sum of connected multigraph types limited by 3 edges and considering partitions of genera"}. It performs a Python calculation to obtain the first experimental data. From theoretical considerations, it correctly identifies the shape of the final answer, writing {\it "Ultimately, I want a final closed formula for $N_g$ as a degree-3 quasi-polynomial with a period of 6."}. After a few attempts, it calculates this polynomial via Lagrange interpolation on datapoints with fixed residue modulo $6$, discovering that the case $g=2$ needs separate treatment. This not only represents a perfect solution to the given problem, but also mirrors precisely the approach of the human question author to solving the problem.
    \item \grokfour 
    obtains an expression for $N_g$ in a single reasoning step, though no further details are available as the \grokfour API does not expose reasoning summaries. The model then uses a Python tool to calculate the first values and the SageMath tool to look up the resulting integer sequence in the OEIS database \cite{oeis}. This being unsuccessful, it submits a very concise sketch of its answer, which is slightly less simple than the formula for $N_g$ above, as it still features a summation over $g-2$ terms.\\
    In a second evaluation, \grokfour uses the bash tool to download textbooks on algebraic graph theory and moduli spaces of curves and convert them to text. Lacking the software tools for the latter, it tries and fails to install new packages on the sandboxed Docker container, receiving an error for attempting to use sudo rights. Finally, it abandons these attempts and just submits a solution that is mostly correct, but has some small errors in one of the terms.
    \item \claudeopus also tries to combine combinatorial arguments with computer calculations in SageMath, but fails to find even the contribution from $2$-vertex graphs, forgetting some topological possibilities for $\Gamma$. One noteworthy pattern is that the model includes very verbose reasoning in the form of comments and static print statements within the SageMath code.
    \item \geminipro starts with a correct calculation of $N_2, N_3, N_4$. However, then it makes the completely unfounded claim that {\it "This implies that $N_g$ is a quadratic polynomial in $g$."}, whereas in reality it is a \emph{cubic} \emph{quasi}-polynomial. It then submits an answer based on that wrong assumption. It does get partial credit in the subquestions for calculating $N_3=9$ correctly.
\end{itemize}

\section{Detailed Tool Descriptions}\label{app:tools}

The evaluation environment for IMProofBench was designed to emulate the computational resources available to research mathematicians when solving complex problems. Rather than restricting models to basic arithmetic operations, we provide access to the same sophisticated mathematical software that researchers routinely use in their work. This approach reflects the reality that modern mathematical research frequently involves computational exploration, symbolic manipulation, and verification of conjectures through extensive calculation.

\subsection{Technical Specifications}

All tools operate within the following constraints to balance computational power with practical limitations:
\begin{itemize}[left=1em,itemsep=0.1em]
    \item \textbf{Timeout}: 15 minutes per tool invocation
    \item \textbf{Memory limit}: 8 GB RAM per execution
    \item \textbf{Environment}: Isolated Docker container running Arch Linux
    \item \textbf{Execution model}: Independent tool calls (no variables persist between calls), but files written to the filesystem remain accessible throughout the evaluation session
\end{itemize}

\subsection{Core Computational Tools}

\subsubsection{Python Environment}

The Python tool provides access to a comprehensive scientific computing environment (Python 3.13.7). This language was chosen for its prevalence in scientific computing and the extensive familiarity that language models demonstrate with its syntax and libraries. The environment includes standard numerical and symbolic computation packages:

\begin{itemize}[left=1em,itemsep=0.1em]
    \item \textbf{Numerical computing}: NumPy, SciPy, pandas
    \item \textbf{Symbolic mathematics}: SymPy, SymEngine
    \item \textbf{Visualization}: Matplotlib (though output is text-based)
    \item \textbf{Graph theory}: NetworkX, igraph, graph-tool
    \item \textbf{Optimization}: CVXPY with multiple backend solvers (GLPK, ECOS, OSQP, SCS, CSDP)
    \item \textbf{Machine learning}: Basic scikit-learn functionality
\end{itemize}

Each Python execution runs independently with no variables or imports preserved between invocations, though files written to disk remain accessible for subsequent tool calls.

\subsubsection{Bash Shell Access}

The bash tool provides command-line access to the evaluation environment, enabling models to leverage specialized mathematical software that operates through command-line interfaces. This tool serves as the gateway to domain-specific mathematical systems detailed in Section~\ref{subsec:specialized}.

\subsubsection{SageMath}

SageMath~\citep{sagemath} (version 10.6) serves as the primary computer algebra system, providing a unified Python-based interface to numerous mathematical software packages. Its significance in the research community stems from its comprehensive coverage of mathematical domains and its philosophy of combining the best open-source mathematics software into a coherent system.

Key features available through the \texttt{sage\_computation} tool include:
\begin{itemize}[left=1em,itemsep=0.1em]
    \item Natural mathematical syntax through automatic preparsing (e.g., \texttt{x\^{}2} for exponentiation, \texttt{K.<a>} for field extensions)
    \item Extensive algebraic capabilities: polynomial rings, number fields, elliptic curves, modular forms
    \item Combinatorial structures: graphs, matroids, posets, designs
    \item Specialized packages: \texttt{admcycles} for moduli spaces of curves, \texttt{ore\_algebra} for D-finite functions and recurrence operators, \texttt{pari\_jupyter} for enhanced PARI/GP integration
    \item Integration with external systems: automatic interfacing with GAP, Maxima, PARI/GP, Singular
\end{itemize}

\subsection{Specialized Mathematical Software}\label{subsec:specialized}

The evaluation environment includes a comprehensive suite of specialized mathematical software, accessible through the bash tool:

\subsubsection{Computer Algebra Systems}

\begin{itemize}[left=1em,itemsep=0.1em]
    \item \textbf{GAP} (Groups, Algorithms, Programming): Specialized system for computational discrete algebra, particularly group theory and combinatorics~\cite{gap}
    \item \textbf{Maxima}: General-purpose computer algebra system for symbolic computation, descended from MIT's Macsyma~\cite{maxima}
    \item \textbf{PARI/GP} (version 2.17.2): High-performance system focused on number theory computations~\cite{pari}
    \item \textbf{Singular}: Specialized system for polynomial computations, commutative algebra, and algebraic geometry~\cite{singular}
    \item \textbf{Polymake} (version 4.14): System for research in polyhedral geometry and related areas~\cite{polymake}
\end{itemize}

\subsubsection{Algebraic and Geometric Computation}

\begin{itemize}[left=1em,itemsep=0.1em]
    \item \textbf{Normaliz}: Computation of normalizations of affine semigroups and rational cones~\cite{normaliz}
    \item \textbf{LattE integrale}: Lattice point enumeration and integration over convex polytopes~\cite{latte}
    \item \textbf{Gfan}: Gröbner fans and tropical varieties computation
    \item \textbf{4ti2}: Algebraic, geometric, and combinatorial problems on linear spaces
    \item \textbf{msolve}: Polynomial system solving over finite fields and rational numbers
\end{itemize}

\subsubsection{Graph Theory and Combinatorics}

\begin{itemize}[left=1em,itemsep=0.1em]
    \item \textbf{nauty and Traces}: Graph automorphism and canonical labeling~\cite{nauty}
    \item \textbf{bliss}: Another efficient graph automorphism tool
    \item \textbf{igraph}: Network analysis and graph algorithms library
\end{itemize}

\subsubsection{Optimization Solvers}

\begin{itemize}[left=1em,itemsep=0.1em]
    \item \textbf{Linear Programming}: GLPK (GNU Linear Programming Kit), Gurobi-compatible interfaces
    \item \textbf{Mixed-Integer Programming}: SCIP (Solving Constraint Integer Programs)~\cite{scip}
    \item \textbf{Semidefinite Programming}: CSDP, DSDP for SDP problems
    \item \textbf{SAT Solvers}: glucose, kissat, cryptominisat for Boolean satisfiability
\end{itemize}

\subsubsection{Proof Assistants and Verification}

\begin{itemize}[left=1em,itemsep=0.1em]
    \item \textbf{Lean}~\citep{lean4}: Interactive theorem prover and functional programming language
    \item \textbf{Mathics}: Open-source alternative to Mathematica for symbolic computation
\end{itemize}

\subsubsection{Numerical and Scientific Computing}

\begin{itemize}[left=1em,itemsep=0.1em]
    \item \textbf{Julia}: High-performance language for numerical computing
    \item \textbf{SciLab}: Numerical computational package similar to MATLAB
    \item \textbf{FLINT}: Fast Library for Number Theory
    \item \textbf{NTL}: High-performance number theory library
\end{itemize}

\subsection{Data Resources}

The environment includes numerous mathematical databases accessible through SageMath:
\begin{itemize}[left=1em,itemsep=0.1em]
    \item Stein-Watkins database of elliptic curves
    \item Jones' database of number fields
    \item Kohel database for elliptic curves and modular polynomials
    \item Cunningham tables for factorizations
    \item OEIS (Online Encyclopedia of Integer Sequences) integration
    \item Various polytope databases and mutation class data
\end{itemize}

\subsection{Web Search Capabilities}

The \texttt{web\_search} tool provides access to current mathematical literature and online resources. The implementation follows a provider-based architecture:

\begin{itemize}[left=1em,itemsep=0.1em]
    \item \textbf{Internal providers}: Models from OpenAI, Anthropic, and Grok utilize their respective built-in web search capabilities, requiring no additional API keys
    \item \textbf{External provider}: Tavily is configured as a fallback for models without internal search capabilities (e.g., Gemini), providing AI-optimized search results
\end{itemize}
Some models, notably \grokfour, combine web search capabilities with the \texttt{wget} bash command to download full research papers for detailed analysis.

\subsection{Example Tool Uses From Benchmark Evaluation} \label{sec:exa_tooluse}
Below, we list some example tool applications that occurred during our model evaluations. In each case, the full log file of the multi-turn evaluation reveals that the respective calculation played a decisive role in allowing the model to find the correct answer. To preserve benchmark privacy, we describe the relevant tool uses in general terms while leaving out the details of the specific benchmark problem.
\begin{itemize}[left=1em,itemsep=0.1em]
    \item \textbf{Generating functions} ({Model}: \grokfour, {Tool}: SageMath)\\ Solved combinatorics problem by calculating a generating function $F(x)$ and forming the exponential $G(x)=\exp(F(x))$ to extract a specific coefficient from $G$ 
    \item \textbf{Modular forms} ({Model}: \grokfour, {Tool}: SageMath)\\ Compute $q$-expansion of the weight $12$ cusp form $\Delta$ 
    \item \textbf{Group theory} ({Model}: \gptfive, {Tool}: \cite{gap} via Bash Shell)\\ Accessed entries of the character table of a sporadic group
    \item \textbf{Literature access} ({Model}: \grokfour, {Tool}: Bash Shell)\\ Model uses \texttt{curl} to download PDF of paper from arXiv, installs the \texttt{PyPDR2} package via \texttt{pip}, and converts the PDF to text to obtain relevant information for the benchmark problem. Note: after an initial failed attempt at installing the \texttt{PyPDF2} package, the model uses the \texttt{pip} argument \texttt{--break-system-packages} to force a user installation in the externally managed Python environment of our sandboxed evaluation environment.
\end{itemize}

\section{Plans for future development}\label{app:development}
Below, we give further details on our plans for the continuous development of IMProofBench.
\begin{itemize}[left=1em,itemsep=0.1em]
\item \textbf{Scale and outreach}: We aim to expand the benchmark to 150--300 problems, e.g.\ through strategic partnerships with leading mathematical institutions (e.g., MFO Oberwolfach, IAS, Fields Institute) and by recruiting domain-specific ambassadors who can promote participation at conferences and within their research networks.

\item \textbf{Quality assurance and grading}: To strengthen the scientific validity of our evaluations, we will study inter-rater reliability by comparing expert gradings on the same problems. We will support graders via AI-assisted pre-screening of model answers and refine our error classification system to localize specific mistakes within solution texts rather than applying only global categories.

\item \textbf{Dynamic problem management}: As mathematical knowledge evolves, problems may become easier due to new publications or techniques. We will implement a generous retirement policy allowing authors to withdraw problems affected by recent research, while regularly adding fresh problems to maintain benchmark difficulty. We also plan to release small sets of sample problems to provide the community with concrete reference points for gauging AI progress.

\item \textbf{Technical innovation}: We plan to develop automated difficulty classifiers to predict which problems challenge current AI systems, explore alternative evaluation formats (such as formula reconstruction tasks and interactive problem-solving sessions), and implement bring-your-own-agent interfaces to enable companies to test internal models against the benchmark.

\item \textbf{Evaluation modalities}: Building on the existing IMProofBench platform and contributor network, we plan to explore further problem types and evaluation methodology. This includes:
\begin{itemize}
    \item combinations of informal and formalized questions and solutions (e.g., in collaboration with the ProofBench project \cite{proofbench}),
    \item specialized task formats with wide importance to mathematical research, such as formula reconstruction for sequence data of natural/rational numbers, polynomials, $\ldots$ (see e.g. \cite{gauthier2023learning, belcak2022fact, pmlr-v162-d-ascoli22a}),
    \item interactive or collaborative proof attempts, including provision of hints or feedback to the model during evaluation time, more closely mimicking the setting of a researcher using commercially available AI systems.
\end{itemize}

\end{itemize}

\section{Use of Large Language Models}
\label{app:ai_usage}

We report our use of LLMs throughout this research project. The authors take full responsibility for all content in this paper, including any LLM-assisted portions.

\subsection{Writing and Presentation}

Claude Opus was used to generate an initial draft of Sections 3 and Appendices \ref{app:interface}, \ref{app:tools}, and \ref{app:ai_usage}. Additionally, Claude Opus provided proofreading assistance and offered stylistic and structural suggestions. 
GPT-5 was used to collect bibliography entries of software packages in Appendix D.
All LLM-generated content was thoroughly reviewed, fact-checked, and edited by the authors.

\subsection{Literature Discovery and Related Work}

During the ideation phase and preparation of the benchmark whitepaper, we used ChatGPT o3 and Claude Opus Research to conduct comprehensive searches of the benchmarking literature and identify related projects. These tools helped surface relevant prior work and ensure thorough coverage of the existing landscape, though all citations were independently verified by the authors.

\subsection{Research Implementation and Development}

LLMs played a substantial role in implementing the benchmark infrastructure. Claude Code, supplemented by ChatGPT's Codex CLI tool, assisted with:
\begin{itemize}[left=1em,itemsep=0.1em]
    \item Development of the benchmark website and database architecture
    \item Adaptation of the Inspect framework for model evaluation
    \item Extraction and visualization of quantitative results
\end{itemize}

Additionally, Claude Opus provided support for organizational tasks, including meeting summaries and the creation of promotional materials. The background image for our benchmark poster was generated using GPT-5's multimodal capabilities.

All code and implementations were tested, validated, and debugged by the authors to ensure correctness and functionality.

\section{Evaluation Prompts}\label{app:prompts}

\begin{prompt}{Main Question Prompt}
# Background
The IMProofBench project is a mathematical reasoning benchmark for AI systems, testing their ability to solve research level math problems. Each such problem consists of one **main question**, where the expected answer is a longform mathematical proof, and several related **subquestions** which have short, unique answers (e.g. a natural number). The main answer will be graded by both human expert mathematicians (often the author of the question) and AI evaluators, whereas subquestion answers are checked automatically using a Python script.

# Structure of the evaluation
In the following we would like to evaluate your mathematical reasoning abilities on one such problem. The overall structure of the conversation below is that we iterate through the questions in order (main question, sub-question 1, sub-question 2, ...) and in each step, you can:

- Read the current question
- Think about it in a multi-turn environment with tool use (see below)
- Submit the answer to the current question

At each point in the conversation, you have the context of the entire previous conversation including your outputs in the thinking steps and the record of any tool uses. Note that you will *not necessarily* have access to records of your internal reasoning traces and internal tool uses, so any helpful information from these should be documented in your (external) thinking outputs. 

# Multi-turn reasoning environment
To help you solve the problem, you will have access to a multi-turn conversation environment with optional tool use, based on the Inspect AI framework. At each step, you can:

- Think out loud to analyze the problem, devise a solution approach, think through the steps of mathematical arguments, etc.
- Use the `python` tool to run self-contained experiments in a standard python environment
- Use the `bash` tool to execute commands inside a docker container (running ArchLinux with some open-source mathematical software installed)
- Use the `web_search` tool to search for current information, mathematical definitions, theorems, or recent research
- Use the `sage_computation` tool for conducting an experiment in a self-contained SageMath terminal session
- Use the `submit` tool to provide your final answer to the current question (main or sub-question)

All tools have a timeout of 15 minutes, maximal memory usage (RAM) of 8 GB and run on standard 2025 hardware.

# Token constraints
You have {main_question_token_limit:,} tokens to solve the main question, and {subquestion_token_limit:,} tokens for each of the following sub-questions. This counts both your output tokens (including in tool calls) and your reasoning tokens. You are informed about your current usage after each conversation turn.

# Answer format for main question
Below you will see the text of the main question. Once you finished reasoning about it, you can register your answer using the `submit` tool. The answer for the main question should be a detailed mathematical argument, formatted in Markdown with LaTeX formulas using $...$ for inline mathematical expressions and $$...$$ for equations. Use Markdown [link formatting](https://www.markdownguide.org/basic-syntax/#links) for including online references, *not* any internal web-referencing system.

**Problem:**

{question_text}

Please work through this problem step by step. When you have your final answer, use the submit() tool to submit it.
\end{prompt}

\begin{prompt}{Subquestion Prompt}
---

**Great work on the previous part!** 

You have successfully completed the previous question. Now please solve the following subquestion while keeping the context of your previous work:

**Subquestion {subquestion_order}:**
{subquestion_text}

**Instructions:**
- You can reference your work from previous parts
- Use the same mathematical tools available to you  
- When you have your final answer, use the submit() tool to submit it
- Be precise and specific in your answer format

Please proceed with solving this subquestion.
\end{prompt}

\begin{prompt}{Conversation Status Update}
Please continue working on the current question. To formally register your answer, use the `submit` tool as per the original instructions above. Note: the conversation will only proceed to the next stage once you use the `submit` tool.

Token usage: {current_tokens:,} of {token_limit:,} tokens used for this stage.
\end{prompt}

\begin{prompt}{Python tool description}
Use the python function to execute Python code.

The Python tool executes single-run Python scripts. Important notes:
1. Each execution is independent - no state is preserved between runs
2. You must explicitly use print() statements to see any output
3. Simply writing expressions (like in notebooks) will not display results
4. The script cannot accept interactive input during execution
5. Return statements alone won't produce visible output
6. All variables and imports are cleared between executions
7. Standard output (via print()) is the only way to see results
8. This tool has a timeout of 15 minutes and maximal memory usage (RAM) of 8 GB
\end{prompt}

\begin{prompt}{Bash tool description}
Use this function to execute bash commands. Underlying system is ArchLinux with many standard open-source computer algebra systems (like GAP) pre-installed.
This tool has a timeout of 15 minutes and maximal memory usage (RAM) of 8 GB.
\end{prompt}

\begin{prompt}{Web search tool description}
Use this function to search the web for current information, mathematical definitions, theorems, or recent research.

This tool gives you access to up-to-date information that can help with:
- Looking up mathematical definitions and theorems
- Finding recent research papers or results
- Verifying computational results against known databases
- Checking current mathematical conventions or notation
- Finding examples of similar problems or techniques

The search results will include titles, URLs, and relevant excerpts from web pages.
Use this tool when you need information that might not be in your training data or when you want to verify facts.
\end{prompt}

\begin{prompt}{Sage tool description}
Use the sage_computation function to run calculations in the open-source mathematics 
software system SageMath.

The sage_computation tool executes single-run SageMath scripts. Important notes:
1. Each execution is independent - no state is preserved between runs
2. You must explicitly use print() statements to see any output
3. Simply writing expressions (like in notebooks) will not display results
4. The script cannot accept interactive input during execution
5. Return statements alone won't produce visible output
6. All variables and imports are cleared between executions
7. Standard output (via print()) is the only way to see results
8. This tool has a timeout of 15 minutes and maximal memory usage (RAM) of 8 GB

All standard SageMath functions are pre-imported and available.
The SageMath preparser is applied, so you can use natural mathematical syntax.

Key Features:
- Natural syntax: Use x^2 for powers, K.<a> for field extensions
- All mathematical objects pre-imported: Matrix, EllipticCurve, PolynomialRing, etc.
- Advanced packages available: admcycles for moduli spaces, and many more

Examples:
  # Factor a polynomial
  factor(x^100 - 1)
  
  # Define a number field
  K.<a> = NumberField(x^3 - 2)
  
  # Work with elliptic curves
  E = EllipticCurve([0, 1])
  print(E.rank())
  
  # Use specialized packages (example with admcycles)
  from admcycles import *
  G = StableGraph([1,1],[[1,3],[2,4]],[(1,2),(3,4)])
  print(f"Automorphisms^2: {G.automorphism_number()^2}")

IMPORTANT: Like the python() tool, you must use print() to see any output.
Nothing is returned automatically - always print your results!
\end{prompt}

\begin{prompt}{Submit tool description}
Submit your final answer for the current question or subquestion. Use Markdown + LaTeX formatting.
The answer for the main question should be a detailed mathematical argument.

Your answer should be formatted as natural Markdown text with LaTeX formulas.
Use $ for inline math and $$ for display math, or \begin{equation} environments.
Use standard [Markdown link syntax](https://www.markdownguide.org/basic-syntax/#links) for online references.

RECOMMENDED: Use raw strings (r''' or r"") to write LaTeX naturally without escaping.

Important formatting notes:
- Write your answer exactly as you would in a math document
- Use raw triple quotes r''' for multiline answers with LaTeX
- This lets you write \frac, \sqrt, \int naturally (no escaping needed)
- Include full mathematical reasoning with the final answer clearly stated
- Do not use custom macros (e.g., \Z, \Q, \RR, etc.). Only use valid standard LaTeX commands
\end{prompt}

\begin{prompt}{Non-agentic prompt for first subquestion}
Thanks! Below I'll ask some follow-up questions, where answers will be parsed automatically. They might repeat the main question or ask for results in special cases. Please provide the final answer in a \boxed{} environment for easier extraction. This desired answer will typically be short, either \boxed{Yes} or \boxed{No} or a numerical result like \boxed{172} or \boxed{-13/45}. Do not use additional (LaTeX) formatting within the \boxed environment.

Subquestion 1:
{subquestion_text}
\end{prompt}

\begin{prompt}{Non-agentic prompt for further subquestions}
Thanks!
Subquestion {number}:
{subquestion_text}
\end{prompt}

\newpage

\end{document}